\documentclass[authoryear, 11pt]{article}

\usepackage{graphicx}
\usepackage{subcaption}
\usepackage{amssymb}
\usepackage{booktabs}
\usepackage{tabularx}
\usepackage[document]{ragged2e}
\usepackage{graphicx}
\usepackage[table,xcdraw]{xcolor}
\usepackage{lscape}
\usepackage{adjustbox}
\usepackage{comment}
\usepackage{graphicx}      
\usepackage{subcaption}    
\usepackage{caption}       

\usepackage[T1]{fontenc}
\usepackage{amsmath} 
\usepackage{tablefootnote}
\usepackage{graphics}
\usepackage{epsfig}

\usepackage{amssymb}

\usepackage{amsthm}
\usepackage[colorlinks = true,
           linkcolor = blue,
            urlcolor  = blue,
            citecolor = blue,
            anchorcolor = blue]{hyperref}
\usepackage{url} 
\usepackage{natbib}
\setcitestyle{aysep={,}} 

\usepackage{ragged2e}
\usepackage{makecell}
\usepackage{authblk} 

\usepackage[left=2cm, right=2cm, top=3cm, bottom=3cm, footskip=0.5cm]{geometry}
\usepackage{adjustbox}
\usepackage{array}
\usepackage{pdflscape}
\usepackage{longtable}

\definecolor{headercolor}{gray}{0.4}
\definecolor{lightgray}{gray}{0.9}

\usepackage{authblk}
\author[1, *]{Grazia Sveva Ascione}
\author[2]{Nicolò Tamagnone}
\affil[1]{Department of Industrial Engineering, Polytechnic University of Turin\rule[30pt]{0pt}{0pt}}
\bigskip
\affil[2]{Venice School of Management, Ca Foscari University of Venice}

\vspace{4em} 
\bigskip
\affil[*]{\textbf{Corresponding author: \texttt{grazia.ascione@polito.it}}
}

\title{\textbf{From scratch to silver: Creating trustworthy training data for patent-SDG classification using Large Language Models}}

\begin{document}

\maketitle

\begin{abstract}

\justifying

Classifying patents by their relevance to the United Nations Sustainable Development Goals (SDGs) is essential for assessing how technological innovation contributes to global challenges. However, the field lacks a gold-standard, large-scale labeled dataset, limiting the application of supervised learning methods. While previous studies have employed keyword search, transfer learning, or citation-based heuristics, none offer a scalable or generalizable solution.
In this work, we treat patent-to-SDG classification as a \textit{weak supervision} (WS) problem, using citations from patents to SDG-tagged scientific publications, the non-patent literature citations (NPL),  as an initial, noisy proxy for SDG relevance. While informative, this signal is both sparse and incomplete, motivating the need for a more structured, denoised labeling strategy. Therefore, we design a composite labeling function (LF) that combines large language models (LLMs) for semantic concept extraction with a retrieval-based alignment strategy. Using a patent ontology, we extract structured abstractions, \emph{functions}, \emph{solutions}, and \emph{applications}, from both patents and SDG publications. We then compute cross-domain similarity scores and fuse them using a rank-based retrieval mechanism. The function is calibrated using a positive-only loss that optimizes alignment with the original NPL-SDG citations while not penalizing the algorithm for discovering additional SDG associations.
The resulting \textit{silver-standard} dataset assigns soft, multi-label SDG relevance scores to patents. 
The dataset significantly expands the coverage of SDG labels across patents and serves as a high-quality signal for training a multi-label regression model. Validation's results show that our approach outperforms keyword-based, transformer-based, and zero-shot LLM baselines in recovering NPL-SDG signal. Additionally, the resulting silver labels produce higher modularity in patent citation, co-inventor, and co-applicant networks, indicating greater thematic, cognitive, and organizational coherence also when compared with classic technological classifications.
Finally, we demonstrate that the silver labels are consistently learnable in a multi-label regression setting, confirming that LLM-derived semantic features encode meaningful signals. These findings highlight the value of weak supervision in correcting and expanding noisy signals, and show that semantically grounded classifications can uncover dimensions of innovation  beyond what traditional technological taxonomies reveal.\end{abstract}

\clearpage

\justifying

\section{Introduction}

\label{sec:intro}

As the world reaches the halfway point toward achieving the United Nations Sustainable Development Goals (SDGs) by 2030, progress remains alarmingly slow. According to the \citet{WIPO2024} SDG Report, only 15 percent of the targets are currently on track, leaving many of the world’s most vulnerable populations at risk. This underscores the urgent need for collective action to accelerate progress.

Innovation and creativity are essential drivers in bridging this gap, with intellectual property (IP) systems playing a crucial role in incentivizing the development and dissemination of new technologies \citep{mazzi2024patents}. By protecting and rewarding inventors, IP facilitates the commercialization of solutions to pressing global challenges, including climate change, public health crises, and clean energy transitions. However, despite the central role of technological innovation in sustainable development, measuring and tracking progress in this domain remains a significant challenge.

Patent data encapsulates nearly 70 \% of global technological knowledge and is largely publicly available \citep{WIPO2024}. Yet, extracting meaningful insights from this vast and complex dataset is not straightforward. Patent analytics provides a solution by transforming raw patent data into actionable intelligence, allowing researchers and policymakers to monitor technological advancements, identify research gaps, and strategically direct investments. By leveraging these insights, stakeholders can enhance research, development, and implementation of technologies that directly contribute to the SDGs.

Existing attempts to map patents to SDGs face a foundational obstacle: the lack of a high-quality, large-scale labeled dataset.  In response, prior research has adopted a range of strategies, including keyword-based filtering \citep{van2021innovative}, transfer learning from SDG-labeled scientific articles \citep{hajikhani2022mapping}, and supervised learning on jurisdiction-specific annotated datasets \citep{maehara2025multi}. Further, another strategy proposed by \citet{ciarli2022changing} uses citations from patents to SDG-tagged scientific articles in the non-patent literature (NPL) as an indirect indicator of relevance. This approach leverages SDG labels from the scientific domain to infer SDG alignment in the technological domain. However, the NPL signal suffers, in general, from two critical limitations: it is \textit{noisy} and \textit{incomplete} \citep{callaert2006traces, he2007evidence, nagaoka2015use}. First, citations in patents are often inserted by examiners rather than inventors, and may reflect legal or procedural considerations rather than substantive conceptual links \citep{alstott2017mapping}. Second, the coverage of NPL citations is uneven across technological domains and jurisdictions, and many patents relevant to the SDGs may not cite any scientific publications at all \citep{ciarli2022changing}. As such, NPL-based labeling cannot serve as a reliable or comprehensive basis for training SDG classification models.

In this research, we propose an alternative path: treating patent-to-SDG classification as a problem of \emph{weak supervision} (WS). Rather than relying on hand-labeled data or heuristic taxonomies, we build a scalable \emph{labeling function} (LF) that  assigns SDG relevance scores based on semantic similarity between patents and SDG-relevant scientific publications. This function operates without any manual annotation and is designed to work in the absence of a gold standard.

Our LF is composed of three integrated components. First, we leverage large language models (LLMs) to extract structured semantic concepts, namely \textit{functions}, \textit{solutions}, and \textit{applications}, from both patents and SDG-labeled papers. These abstractions, informed by a patent ontology \citep{zhai2022patent}, provide a shared conceptual space in which scientific and technological texts can be compared. Second, we compute semantic similarity across concept types using domain-specific embeddings, and fuse the resulting rankings using Reciprocal Rank Fusion (RRF). Third, we apply a weakly supervised optimization procedure to calibrate threshold and ranking parameters, aligning our system with known positive anchors from NPL citation data while avoiding overfitting to noisy or incomplete signals. In particular, the loss function used in the optimization phase, optimizes the algorithm to recover the signal from NPL but does not penalize it if it finds new signal, expanding the original one from NPL.
The output of this pipeline is a large-scale, \textit{silver-standard} dataset that assigns soft SDG relevance vectors to patents. These labels are then used to train several downstream regression models that predicts SDG alignment directly from patent abstracts and titles. 

Despite being an unsupervised method, our LF consistently outperforms a range of both supervised and unsupervised baselines. In the first \textit{internal }evaluation, it achieves higher recall against NPL-derived SDG signals than classic retrieval methods like \texttt{BM25}, fine-tuned transformers such as \texttt{BERT-for-Patents}, and domain-adapted architectures like \texttt{PAECTER} and \texttt{PAT-SPECTER} \citep{ghosh2024paecter}. It also surpasses zero-shot prompting with \texttt{GPT-4.1}, an emerging baseline for multi-label classification. Crucially, the method not only recovers known NPL-derived associations, but also assigns SDG labels to patents with no citation links to the scientific literature. In the second \textit{external} evaluation, we assess the structural validity of these additional labels by measuring overlapping modularity \citep{nicosia2009extending} in three networks: patent citations, co-inventors, and co-applicants. In all cases, our method outperforms both NPL-based labels and CPC classifications in aligning with citation, inventor, and applicant networks. While CPC reflects technological domains rather than sustainability goals, this result suggests that a semantically grounded SDG classification can capture thematic and organizational coherence in innovation that may not be fully reflected by technology-based taxonomies.
While our approach demonstrates valuable empirical performance, several limitations remain. The LF still depends on the quality of the external ontology and the coverage of NPL citations. The absence of gold-standard labels limits the evaluation of soft, multi-label predictions. Additionally, reliance on LLMs introduces potential issues such as semantic drift and domain sensitivity. 

The remainder of this paper is structured as follows. Section~\ref{sec:literature} reviews related work on patent–SDG classification and weak supervision in the patent literature. Section~\ref{sec:datamet} describes the dataset construction and the design of the labeling function, including semantic abstraction, ranking step and weakly supervised optimization. Section~\ref{sec:valid} provides  validation for the proposed method. Section~\ref{sec:res} presents the resulting silver-standard dataset and evaluates its performance in a multi-label classification setting. Section \ref{sec:limitations} presents the main limitations while Section~\ref{sec:conclusion} concludes.

\clearpage
\section{Literature review}
\label{sec:literature}

\subsection{Mapping patents to non-technological domains: the case of SDGs}

A standard machine learning (ML) task in innovation studies is the classification of patents based on their technological content, typically using hierarchical taxonomies such as the International Patent Classification (IPC) or the Cooperative Patent Classification (CPC) \citep{li2018deeppatent, lee2019patentbert, haghighian2022patentnet}. These taxonomies provide a structured way to categorize patents into technological domains and support the development of supervised classification systems.

More recently, there has been growing interest in classifying patents into domains not explicitly covered by existing taxonomies, particularly in areas with policy relevance or interdisciplinary characteristics. For example, the classification of patents related to critical raw materials (CRMs) often requires combining CPC codes with keyword-based or semantic techniques \citep{de2025mapping}. Similarly, the circular economy represents a broader and evolving classification challenge. While some dedicated CPC codes exist, researchers frequently rely on hybrid strategies that combine keyword-based named entity recognition with CPC filters \citep{giordano2024identifying, caldarola2024economic}.

Mapping patents to the United Nations' Sustainable Development Goals (SDGs) poses an even more complex task. Unlike circular economy domains, the SDGs are not defined by technological areas but by societal outcomes, including poverty reduction, health, education, and environmental sustainability. This makes the classification problem more abstract and less directly linked to technological taxonomies. Furthermore, the SDGs are not mutually exclusive: a single patent may contribute to multiple goals simultaneously, which challenges single-label classification frameworks and motivates the use of multi-label or probabilistic approaches.

To address these challenges, several methodological approaches have been proposed. A widely used strategy is keyword-based classification. For instance, \citet{van2021innovative} developed a thesaurus of SDG-related keywords based on Agenda 2030 and differentiated between "blue patents" (focused on unmet social needs) and "green patents" (centered on established sustainable technologies). \citet{ascione2023technological} extended this approach by combining TF–IDF with vector-based text representations to expand the vocabulary of SDG-related terms. However, keyword methods suffer from significant limitations, including lack of contextual understanding, synonym mismatches, and susceptibility to polysemy. These methods also require intensive manual curation. While some expert-reviewed keyword sets exist \citep{bordignon2021dataset, kashnitsky2022identifying}, no widely accepted SDG-specific patent dictionary is currently available.

These limitations are evident in cases where keywords yield false positives. For instance, the term \emph{rural}, used by \citet{van2021innovative} to identify blue patents, may match patents unrelated to sustainable development, such as CN109446992B, which describes a remote sensing AI system potentially used in military or commercial real estate applications.\footnote{The text of the cited patent is available at \url{https://patents.google.com/patent/CN109446992B/en?q=(rural)&num=100&oq=rural}.}

Beyond keyword-based methods, ML techniques have been applied to SDG patent classification. \citet{hajikhani2022mapping} used transfer learning by training a model on SDG-labeled scientific publications and applying it to European patents. While this method enables domain transfer, its effectiveness is limited by the linguistic and structural differences between scientific literature and patent texts, and by the assumption that each patent maps to only one SDG.

A more advanced supervised method was proposed by \citet{maehara2025multi}, who trained a BERT-based classifier (DEBERTA) on a dataset of 8,000 Japanese patents manually labeled by the Japan Patent Office (JPO). This approach demonstrates strong performance but depends on jurisdiction-specific data and costly expert annotation, which may limit its generalizability.

An alternative strategy was introduced by \citet{ciarli2022changing}, who proposed tagging patents with SDGs based on non-patent literature (NPL) citations. Specifically, patents that cite SDG-related scientific articles are labeled with the corresponding goal. Although intuitive, this method is subject to important limitations. Citations may be noisy or strategic rather than semantically relevant \citep{lampe2012strategic, kuhn2020patent}, and many patents relevant to SDGs may not contain NPL citations at all.

Table~\ref{tab:sdg_mapping_approaches} provides a comparative overview of the main approaches found in the literature, highlighting their methodological characteristics and key limitations.

\begin{table}[htbp]
\centering
\caption{Overview of main approaches to mapping patents to SDGs}
\label{tab:sdg_mapping_approaches}
\begin{tabularx}{\textwidth}{>{\hsize=0.18\hsize}X >{\hsize=0.20\hsize}X >{\hsize=0.31\hsize}X >{\hsize=0.31\hsize}X}
\toprule
\textbf{Approach} & \textbf{Study} & \textbf{Key Features} & \textbf{Limitations} \\
\midrule

\textbf{Keyword-based classification} 
& \citet{van2021innovative}, \citet{ascione2023technological} 
& Uses SDG-related keywords, thesauri from Agenda 2030; some methods enhance with TF–IDF or embeddings 
& High manual effort, lacks contextual understanding, polysemy and synonym mismatch issues \\

\addlinespace

\textbf{Transfer learning} 
& \citet{hajikhani2022mapping} 
& Pretraining on SDG-labeled scientific papers; applied to patents via domain transfer 
& Scientific and patent language differ; assumes one SDG per patent; limited validation on patents \\

\addlinespace

\textbf{Supervised classification} 
& \citet{maehara2025multi} 
& Fine-tuned transformer on 8,000 SDG-labeled Japanese patents; high performance 
& Data limited to Japanese jurisdiction; unclear generalizability; expensive labeling \\

\addlinespace

\textbf{Citation-based} 
& \citet{ciarli2022changing} 
& Labels inferred through citations to SDG-labeled publications; aligns patents with scientific content 
& Citations may be noisy or strategic; not all relevant patents have NPL citations \\

\bottomrule
\end{tabularx}
\scriptsize
\justifying\emph{Notes:} The table summarizes the main methods found in the literature for mapping patents to the SDGs, with representative studies, key features, and known limitations.
\end{table}

Despite the progress outlined above, a critical issue remains: the lack of a publicly available and universally accepted \emph{gold standard} for benchmarking SDG classification methods. Most existing approaches rely on task-specific, jurisdiction-dependent, or indirectly inferred labels, which limits comparability across methods and undermines reproducibility. This gap continues to be a barrier for developing scalable, generalizable frameworks for SDG mapping in patent analytics.

\subsection{The annotation bottleneck: Paradigms under limited supervision}

\subsubsection*{Core methodologies}

\begin{table}[htbp]
\centering
\caption{Comparison of limited-supervision learning paradigms in patent classification}
\label{tab:limited_supervision_methods}
\begin{tabularx}{\textwidth}{>{\hsize=0.18\hsize}X >{\hsize=0.20\hsize}X >{\hsize=0.31\hsize}X >{\hsize=0.31\hsize}X}
\toprule
\textbf{Approach} & \textbf{Study (patent)} & \textbf{Key Features} & \textbf{Limitations} \\
\midrule

\textbf{Transfer learning} 
& \citet{voskuil2021improving}, \citet{bekamiri2024patentsberta},
\citet{hajikhani2022mapping}
& Fine-tuning of language models (e.g., BERT) pre-trained on general corpora directly on patent text with or without domain adaptation 
& Requires task-specific labels; performance depends on labeled data availability or validation not possible \\

\addlinespace

\textbf{Semi-supervised learning} 
& \citet{chen2015semi}, \citet{huang2020semi} 
& Combines small labeled sets with large unlabeled corpora using pseudo-labeling or co-training 
& Needs at least some gold-standard labels; label noise can accumulate \\

\addlinespace

\textbf{Active learning} 
& \citet{zhang2014interactive}, \citet{xiong2025scalable} 
& Iteratively selects the most informative data to improve performance
& Still depends on human-in-the-loop annotation \\

\addlinespace

\textbf{Zero-/Few-shot learning} 
& \citet{edwards2024language} 
& Uses prompting or in-context examples for classification without fine-tuning; no task-specific labels 
& Limited generalization on domain-specific tasks; impossible to validate without gold standard \\

\addlinespace

\textbf{Weak supervision} 
& \citet{zhong2023reactie} 
& Uses heuristics and metadata, to generate silver labels; scalable labeling 
& Labels are noisy; requires calibration/denoising to achieve training-quality performance \\

\bottomrule
\end{tabularx}
\scriptsize
\justifying\emph{Notes:} The table compares limited-supervision methods explored in the literature. It highlights their core mechanisms, example applications, and limitations.
\end{table}

Scenarios where gold-standard labels are scarce or even unavailable are common across different domains such as healthcare \citep{halpern2016clinical, reitsma2009review}, natural language generation \citep{xu2024benchmarking} and social sciences \citep{bernhard2025beyond}.
The surge of deep learning exacerbated this issue, as it requires substantial labeled examples, and evolving target definitions can further force continuous re-labeling \citep{Ratner_2017}. While large organizations may absorb annotation efforts, most researchers turn to alternative paradigms to reduce burdens to obtain labeled datasets, such as transfer learning, semi‑supervised learning, active learning, few‑shot and zero‑shot learning and weak supervision. \\
\emph{Transfer learning} adapts models pre‑trained on large-scale data to new tasks, saving labeling efforts. In the patent context, it has recently been adopted in a cross domain scenario without domain-adaptation \citep{hajikhani2022mapping}; however, most work in the patent natural language processing (NLP) space uses transfer learning via fine-tuning pretrained language models directly on patent text, leveraging the general language understanding of models like BERT to excel in patent-specific tasks \citep{bekamiri2024patentsberta, voskuil2021improving}. \\
\emph{Semi‑supervised learning (SSL)} leverages small labeled sets with larger unlabeled corpora, via methods such as pseudo‑labeling or co‑training, to build better classifiers \citep{zhu2005semi}. This technique has been more explored on patent datasets. For instance, \citet{chen2015semi} use co-training, a semi-supervised learning technique, to allow patent annotation even when only a small labeled dataset is available. \citet{kim2024technology} incorporate in fact human defined labels in zero-shot topic labels proposed by BERTopic \citep{grootendorst2022bertopic} in a semi-supervised fashion. \citet{huang2020semi} propose the TRIZ-ESSL (Enhanced Semi-Supervised Learning for TRIZ) using both labeled and unlabeled data to improve the prediction performance in patent classification. \\
\emph{Active learning} is also a main approach to learning with limited labeled data. It tries to reduce
the human efforts on data annotation by actively querying the most important examples, treating  humans as oracles to annotate unlabeled data \citep{settles2009active}.
\citet{zhang2014interactive}, for instance, uses active learning as a way to  efficiently build a patent classifier for electric automobiles using limited labeled data, by prioritizing the most informative patents for manual annotation. Morevoer, the study of   \citet{xiong2025scalable} develops an LLM-assisted active learning framework for multi-label patent classification. By leveraging GPT‑4's uncertainty and semantically informed querying within an iterative human-in-the-loop process, they achieve a 15\% increase in Macro‑F1 and reduce manual annotation by approximately 60\%, using a domain-specific dataset of 100,000 drone patents. Their work proves that implementing active learning in zero-shot LLM classification can lead to significant performance improvements. \\
At the same time, unlike traditional machine learning approaches that rely heavily on manual feature engineering, LLMs benefit from pre-training on massive text corpora, enabling them to learn rich, contextual representations of language. In low or zero-label scenarios, LLMs offer two key strategies: zero-shot learning and few-shot learning. In zero-shot learning, models perform classification tasks without any labeled examples, relying solely on natural language prompts and their internal knowledge. Few-shot learning, on the other hand, enables models to generalize from only a few labeled instances per class, often implemented through in-context learning (ICL) or meta-learning paradigms that rapidly adapt to new tasks with minimal supervision. Despite the promise of these approaches, recent empirical evidence presents a more nuanced picture. For example, \citet{edwards2024language} conducted a large-scale evaluation across 16 text classification datasets, comparing zero- and one-shot ICL with traditional fine-tuning. Their results show that while instruction-tuned LLMs perform competitively in extremely low-data settings, fine-tuned models consistently outperform ICL in high-resource scenarios, particularly in multiclass and multilabel classification tasks. These findings suggest that although LLMs are valuable tools for rapid prototyping or bootstrapping classifiers when annotated data is scarce, fine-tuning smaller, task-specific models remains the most reliable choice when sufficient labeled data is available.\\
Finally, \emph{Weak supervision (WS)} refers to slightly different scenarios where labels exist, but they are noisy, uncertain, imprecise, or incomplete. The goal is to improve learning despite label quality issues, often by combining multiple weak signals to approximate better supervision. To this end, WS employs labeling functions (LF), heuristics, domain rules, metadata cues, and similarity metrics to generate noisy annotations, which are aggregated using probabilistic models  to form a silver-standard dataset \citep{Ratner_2017}. For instance, recent work by \citet{zhong2023reactie} uses WS to avoid costly manual annotation. In their work, the authors introduce \textit{REACTIE} a method for extracting structured chemical reaction information from unstructured scientific text.  Specifically, they leverage distant supervision from patent data by taking structured chemical reactions found in patent documents and automatically converting them into weakly labeled textual examples, achieves substantial improvements and outperforms all existing baselines.

\subsubsection*{Why weak supervision for patent-to-SDG classification?}

As summarized in Table~\ref{tab:limited_supervision_methods}, several paradigms exist to address classification problems under limited supervision. However, the challenge of mapping patents to the SDGs presents a particularly constrained case as SDGs are not grounded in technological categories, and no comprehensive gold standard exists for training or benchmarking. This absence of authoritative labels severely restricts the applicability of standard supervised learning approaches.

Under such conditions, one might consider adopting alternative methods that work under limited supervision, as presented above. However, many of these methods presented encounters significant limitations when applied to SDG classification in the patent domain. Transfer learning approaches rely on the availability of source tasks with high-quality annotations. In this context, scientific articles labeled for SDGs may serve as source data, but prior work has shown that the linguistic and structural differences between scientific texts and patents can undermine model generalization when transferred without domain adaptation \citep{hajikhani2022mapping} and are difficult to validate. Semi-supervised learning, while potentially useful, requires an initial seed set of clean labeled examples to propagate labels and in the SDG case, no such clean seed data is widely available.
Moreover, active learning offers another promising direction but remains labor-intensive. It is particularly effective when human experts can iteratively refine the model by labeling the most uncertain examples. Yet, SDG labeling requires high-level interpretive judgment that crosses disciplinary boundaries, making rapid manual annotation difficult to scale. Similarly, zero- and few-shot learning using LLMs offers rapid prototyping potential, but recent evaluations show that these models tend to underperform on domain-specific, multilabel classification tasks when compared to fine-tuned baselines \citep{edwards2024language}. More importantly, in the absence of a gold standard, it becomes impossible to properly evaluate the outputs of zero-shot systems, leaving their effectiveness unverified.
Moreover, few-shot learning, despite being a more performing alternative than zero-shot, would be virtually impossible considering the variety of scenarios covered by each SDG.  \\

In contrast, weak supervision emerges as a more suitable framework for this problem. Unlike the other paradigms, weak supervision is specifically designed for scenarios in which labeling signals exists, but are partial, noisy, or indirect. This is consistent with the situation encountered when we look at NPL citations of SDG scientific articles. On the one hand, NPL are recognized as meaningful proxies for science–technology linkages \citep{callaert2006traces, ciarli2022changing}. On the other hand, a growing body of research illustrates their limitations. For instance, \citet{he2007evidence} demonstrate systematic noise in NPL references, while \citet{nagaoka2015use} define NPL as \emph{"Not only noisy but highly incomplete"}. The point is corroborated  \citet{narin1997increasing} that state NPL citations provide only an imperfect but useful proxy of knowledge transfer. They are often inserted by examiners, may serve legal purposes, and therefore do not always reflect conceptual proximity between patents and scientific knowledge.

Therefore, while patents citing SDG-related NPL cannot be considered definitive ground-truth for both their noisiness and incompleteness, they may serve as \emph{informative anchors} within a weak supervision framework. Collectively, these observations justify treating such citations as noisy and incomplete, yet semantically meaningful, signals that can guide the construction of silver-standard training data.

\section{Data and methods}\label{sec:datamet}

\subsection{Data}\label{sec:datasel}

In this Section, we present all the sources and the construction method for the dataset that we later use in the WS pipeline. It is composed by three foundational data sources: (i) a corpus of scientific publications tagged with the United Nations SDGs, (ii) a dataset capturing patent-to-paper (P2P) citations, and (iii) a corpus of global patent documents. These sources are systematically integrated to construct the unified analytical corpus, referred to as the \textit{Patent-SDG Dataset}, which constitutes the input to the pipeline.

\begin{figure*}[h]
\centering
\includegraphics[scale=0.25]{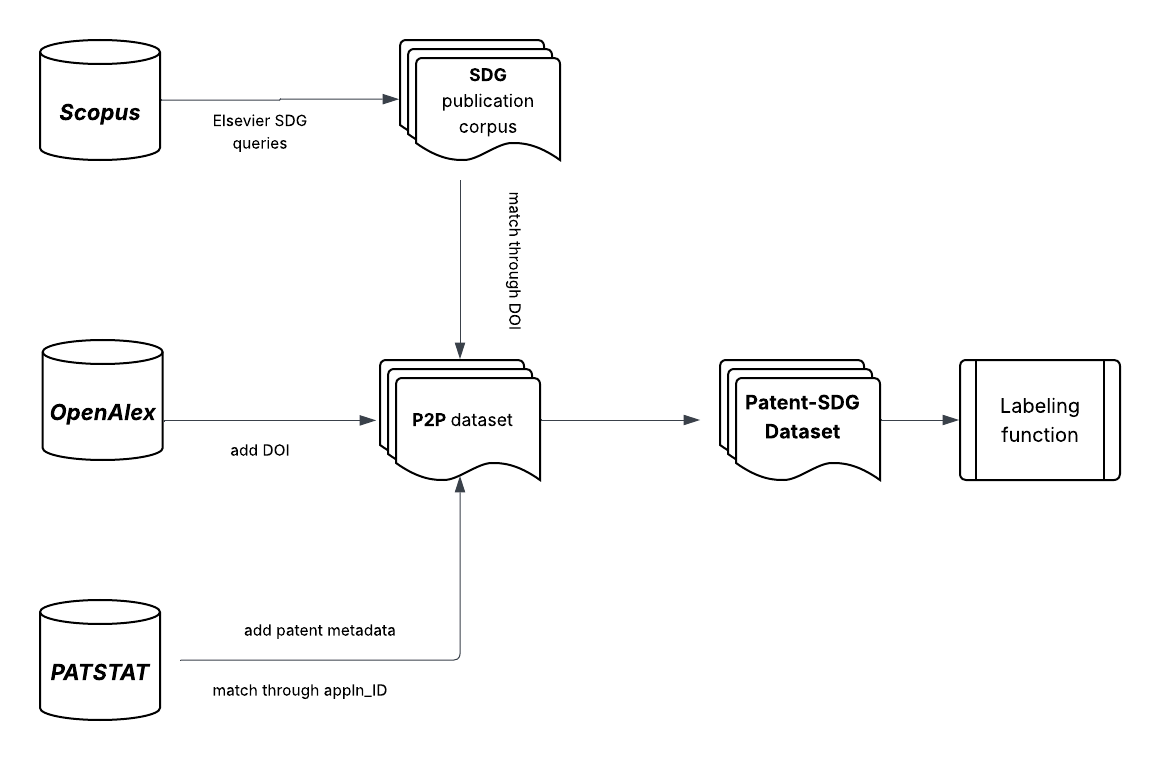}
\caption{Data integration pipeline for constructing the \emph{Patent-SDG Dataset}}
\label{sdgdata}
\scriptsize
\justifying\emph{Notes:} The Figure depicts the data collection and integration pipeline. It starts with three primary data sources: Scopus, used to extract SDG-tagged scientific publications; OpenAlex, which provides DOI-based linking for cited scientific literature; and PATSTAT, used to enrich patent metadata. These components converge through the patent-to-paper (P2P) citation dataset, resulting in the \emph{Patent-SDG Dataset}, which forms the input for the LF and the downstream SDG classification task.
\end{figure*}

Figure~\ref{sdgdata} illustrates the data integration pipeline for the creation of the \textit{Patent-SDG Dataset}, which draws on three major bibliometric and patent data infrastructures: \textit{Scopus}, \textit{OpenAlex}, and \textit{PATSTAT}. Each plays a distinct role in enriching and filtering the P2P citation dataset, described in \citet{marx2020reliance, marx2022reliance}.

First, we employ \textit{Scopus} to construct a corpus of SDG-relevant scientific publications. This is achieved by executing the curated set of queries released through Elsevier’s \textit{SDG Research Mapping Initiative}. These queries enable consistent tagging of publications to one or more of the 17 SDGs.

Second, to link scientific articles cited in patents through NPL citations to their corresponding SDG tags, we integrate data from \textit{OpenAlex}. Specifically, we use OpenAlex identifiers to retrieve DOIs for the cited publications, enabling a reliable join with the SDG-tagged corpus retrieved from Scopus.\footnote{We follow this strategy as in the dataset of \citet{marx2020reliance, marx2022reliance} the only identifier available for each paper is the \textit{openalexid-oaid}, therefore we need to collect the DOI to match to Scopus data.}

Third, we enrich the metadata of the citing patents using the \textit{PATSTAT} database maintained by the EPO. This step allows us to retrieve relevant fields (e.g., title, abstract, filing authority) for patents that reference SDG-related academic literature.

The resulting output, called the \emph{Patent-SDG Dataset}, combines publication-level SDG tagging with citation relationships and textual metadata of patents. This dataset serves as the primary input for the NLP-based classification pipeline developed in this paper.

The following paragraphs give more details on each of the sources.

\vspace{1em}
\noindent\textit{SDG-Tagged Scientific Publications (SDG Corpus).} The SDG corpus comprises scholarly publications associated with one or more of the United Nations SDGs, as identified by Elsevier’s \textit{SDG Research Mapping Initiative}. Since 2018, Elsevier has maintained and periodically updated a set of structured queries for mapping academic output to the 17 SDGs. These mappings are widely adopted, notably within the \textit{Times Higher Education (THE) Impact Rankings}, as indicators of societal impact beyond citation metrics. The latest version of the query set, released in 2023, is publicly available via Elsevier’s data repository \citep{bedard2023elsevier}. The underlying methodology has been validated in several bibliometric analyses \citep{kashnitsky2024evaluating}.

We apply these queries to the \textit{Scopus} database using the Elsevier API\footnote{We accessed the data via the Scopus API. See documentation at \nolinkurl{https://dev.elsevier.com/guides/Scopus\%20API\%20Guide_V1_20230907.pdf}} retrieving English-language journal articles published between 2015 and 2022. Since SDG labels are not natively exposed in the API metadata, we replicate Elsevier's classification using publicly available query definitions.\footnote{Available at: \url{https://elsevier.digitalcommonsdata.com/datasets/y2zyy9vwzy/1}} Given the API's query-length constraints, we develop a method to segment the queries into manageable subqueries, as further detailed in Appendix~\ref{a1}.

The resulting dataset comprises 4,571,753 unique publications. Figure~\ref{sdgpub} in Appendix \ref{a1b} depicts the annual distribution of SDG-tagged articles. A clear upward trend is observable over time, with SDG~3 (Good Health and Well-being), SDG~4 (Quality Education), and SDG~7 (Affordable and Clean Energy) being the most represented goals.

\vspace{1em}
\noindent\textit{Patent-to-Paper (P2P) Citation Dataset.} This dataset originates from an open-access initiative that links non-patent literature (NPL) citations to global patent filings \citep{marx2020reliance, marx2022reliance}.\footnote{The corresponding dataset is available at: \url{https://relianceonscience.org/patent-to-paper-citations}} It captures both \textit{front-page} citations, which are generally added by examiners and hold legal weight, and \textit{in-text} citations, often provided by inventors to reference scientific sources directly influencing the innovation \citep{bryan2020text}. Both citation types can be interpreted as \textit{weak indicators} of conceptual linkage between science and technology. This treatment aligns with the broader literature on science–technology connections, which views NPL citations as a proxy for the scientific foundations of technological innovation \citep{callaert2006traces, narin1997increasing, ke2020analysis}.The dataset includes 7,096,443 patents  globally filed citing 7,447,610 distinct academic articles. To enrich the dataset, we extract the DOI of each cited paper and retrieve full metadata from \textit{Scopus}.

\vspace{1em}
\noindent\textit{Patent Corpus.} For the patent corpus, we rely on European Patent Office (EPO) data, extracted from \emph{PATSTAT}. In particular, we retrieve
the title and abstract of patents in the P2P dataset. We select titles and abstracts as several studies have demonstrated that these sections are effective for comparing patents to scientific papers and for measuring thematic similarity or knowledge flow between science and technology \citep{magerman2015does, motohashi2024measuring, wang2024trace}. Further, we retrieve other patent metadata such as CPC classes, inventors' and citations' information.

\vspace{1em}
\noindent\textit{Patent-SDG Dataset.} The \emph{Patent-SDG Dataset} is a subset of the patent-to-paper citation dataset, filtered to include patents citing highly-cited SDG publications. For each SDG, we identify the top 20,000 most-cited papers (2015–2022), resulting in a pool of 340,000 publications, of which 253,857 are uniquely classified and 86,143 are associated with multiple SDGs.\footnote{This strategy prioritizes citation impact, which may in fact under represent newer or less frequently cited publications. The authors acknowledge this limitation.}
We identify  55,008 patents, whose application is published from 2015 to 2023, that cite 21,517 of these papers. By expanding  the data to include all citations of SDG-tagged articles, we build a Patent-SDG Dataset in which these patents collectively cite 613,853 papers in total, out of which 36,942 are mapped to at least one SDG. Approximately 10\% of these SDG papers are co-classified under multiple goals. A patent is defined as SDG-relevant if it cites at least one SDG-tagged publication. Table \ref{tab:sdgdiv} summarizes the number of unique SDG-tagged publications and associated patents per goal. The SDGs with the greatest numbers of patents associated are SDG3, SDG7 and SGG9, consistently with what found by previous literature \citep{ciarli2022changing}.

\begin{table}[htbp]
\centering
\caption{Number of unique papers and patents per Sustainable Development Goal (SDG) in the Patent-SDG Dataset}
\label{tab:sdgdiv}
\begin{tabular}{ccc}
\toprule
\textbf{SDG} & \textbf{No. of publications} & \textbf{No. of patents} \\
\midrule
1  & 84     & 212   \\
2  & 1,009  & 1,962 \\
3  & 12,777 & 32,529 \\
4  & 266    & 450   \\
5  & 402    & 909   \\
6  & 1,830  & 2,058 \\
7  & 10,767 & 10,258 \\
8  & 626    & 1,265 \\
9  & 4,724  & 9,127 \\
10 & 708    & 1,535 \\
11 & 1,118  & 2,260 \\
12 & 1,455  & 2,450 \\
13 & 2,745  & 2,875 \\
14 & 1,822  & 2,211 \\
15 & 570    & 974   \\
16 & 256    & 562   \\
17 & 577    & 1,629 \\
\bottomrule
\end{tabular}

\scriptsize
\justifying\emph{Notes:} The Table presents the number of unique publications associated with each SDG (\textit{No. of publications}) and the corresponding number of patents that cite them (\textit{No. of patents}). A patent is associated with an SDG if it cites at least one SDG-tagged publication. Since both papers and patents can relate to multiple SDGs, totals exceed the number of unique items. The dataset includes 41,736 SDG-labeled publications and 73,266 patents.
\end{table}

\clearpage
\subsection{Methods}

This Section outlines the two main components of our methodology for mapping patents to SDGs. First, we describe the construction of a labeling function (LF) within a weak supervision framework, designed to generate soft, multi-label SDG relevance scores for patents by aligning structured semantic concepts with SDG-tagged scientific literature. This process yields a silver-standard dataset of SDG labels that serves as training data (Section \ref{sec:lf}). Second, we present the downstream supervised learning model, which is trained on the silver labels to predict SDG relevance directly from raw patent text using a multi-output regression architecture (Section \ref{sec:class}). 

\subsubsection{Creating a labeling function for patent-SDG mapping task via weak supervision}\label{sec:lf}

In the WS framework, a core idea is to replace costly manual annotation with programmatically defined labeling functions (LFs) \citep{Ratner_2017, cohen2019interactive, denham2022witan}. These LFs encode heuristic, noisy, or incomplete knowledge to generate labels that can be used to train downstream machine learning models. As formalized by \citet{Ratner_2017}, labeling functions are modular components that emit noisy labels for unlabeled data based on domain-specific logic, patterns, or external signals. Importantly, the goal is not for any individual LF to be perfect, but for their collective output to approximate useful training data.
In our case, we define a single, composite labeling function 
\(\lambda(p)\), where \(p\) is a patent document, that outputs a soft, normalized 17-dimensional vector representing the relevance of each SDG to that patent. This vector serves as a \textit{silver label}, an imperfect but informative supervision signal derived without manual annotation that can be used for the downstream patent-SDG classification task.
Despite in conventional weak supervision architecture more LFs may be  defined and combined \citep{Ratner_2017}, our method relies on a single composite LF on as in this specific setting we do not have access to different sources of imperfect labels. While this choice limits the diversity of WS signals, we find that the structured and interpretable nature of our labeling function offers a practical and scalable alternative in the absence of task-specific gold-standard data.
Formally, the proposed labeling function is composed of the following structured components:

\begin{figure*}[h]
\centering
\includegraphics[scale=0.15]{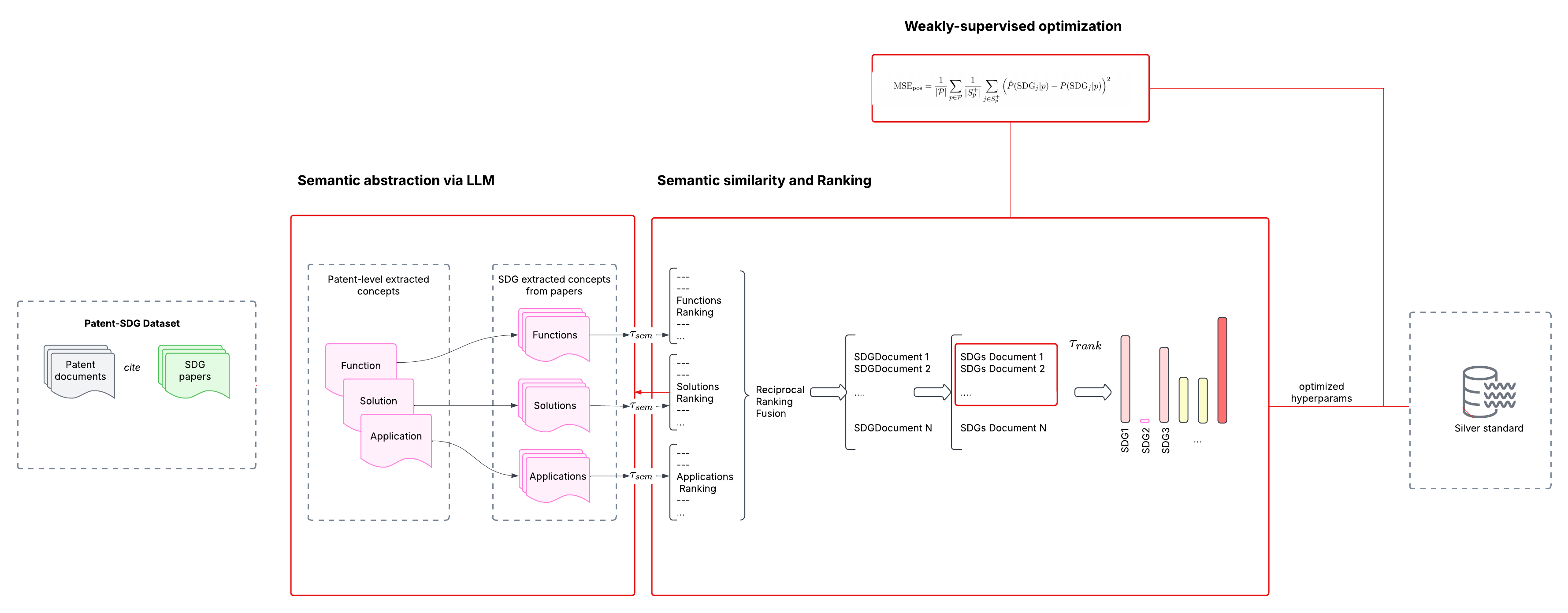}
\caption{Labeling function}
\label{fig:label}
\scriptsize
\justifying{\emph{Notes:}} The Figure illustrates the labeling function pipeline. Patent and SDG documents from the \emph{Patent-SDG Dataset} are processed using LLMs to extract ontological concepts (\textit{functions}, \textit{solutions}, \textit{applications}). Semantic similarities are computed between patent and SDG concepts, and low-similarity pairs (below $\tau_{\text{sem}}$) are filtered out. Concepts are ranked and fused via Reciprocal Rank Fusion (RRF). Weak supervision is applied to optimize fusion weights using SDG labels from cited scientific literature. Final SDG relevance scores (above $\tau_{\text{rank}}$) are used to create the silver standard dataset.

\end{figure*}

\begin{enumerate}
    \item \textit{Semantic abstraction via LLMs}: For each patent \(p\), and each SDG-labeled scientific document \(d\), we extract structured concepts, namely \textit{functions}, \textit{solutions}, and \textit{applications} using a LLM. These abstract concepts form a shared semantic space between scientific articles and patents for alignment.

    \item \textit{Semantic similarity and ranking}: Each concept extracted from the patent is compared to the corresponding concept type in the SDG documents using a cosine similarity over contextual embeddings. For each concept type \(i\), we generate a similarity score \(S_i(p, d)\) and apply a threshold \(\tau_i\) to retain only high-confidence alignments. We then fuse the resulting rankings via Reciprocal Rank Fusion (RRF), producing a final ranked list of SDG documents for each patent.

    \item \textit{Weakly-supervised optimization}: To calibrate the labeling function, we frame the construction of \(\lambda(p)\) as a weak supervision task and optimize the hyperparameters \(\{\tau_i\}, N\) by minimizing a positive-only Mean Squared Error (MSE) loss.
    
    \end{enumerate}

The output of the labeling function \(\lambda(p)\), a soft, 17-dimensional SDG relevance vector for each input patent, constitutes a \textit{silver-labeled dataset}. This dataset serves as the supervisory signal for training a downstream classifier. Specifically, we train neural multi-output regression models that learn to predict the SDG distribution directly from raw patent texts (titles and abstracts). Unlike traditional multi-label classification frameworks that rely on binary or mutually exclusive labels, our regression-based approach preserves the \emph{continuous}, \emph{non-exclusive} nature of SDG alignment. This enables the model to learn nuanced associations and reflect varying degrees of relevance across multiple SDGs simultaneously, in line with the soft supervision provided by our labeling function.

In the following paragraphs, we provide a detailed description of each component of the proposed labeling function, corresponding to the three conceptual blocks outlined in Figure~\ref{fig:label}, including semantic abstraction, similarity-based ranking, and weakly supervised optimization.

\subsubsection*{Semantic abstraction via LLMs} \label{sec:abstraction}

The first step of the LF is to establish a meaningful correspondence between the corpus of SDG documents and the patents' texts of the \emph{Patent-SDG Dataset}. A straightforward solution might involve computing textual similarity between each patent and each SDG documents. This method has been widely adopted in prior studies aiming to link different kinds of documents to patents through textual comparisons \citep{block2021semantic,shibata2011detecting, lippert2024patent}. Moreover, recent research discussed the capabilities of different embedding models to meaningfully capture this relationship without agreement. For instance, \citet{masclans2024measuring} fine-tune \texttt{SCIBERT} \citep{beltagy2019scibert} to predict whether a scientific article will be cited in the NPL of a patent. Their experiments show that \texttt{SCIBERT} outperforms \texttt{SPECTER2} in this task, highlighting the benefits of domain-specific fine-tuning \citep{cohan2020specter, singh2022scirepeval}. Conversely, \citet{ghosh2024paecter} introduce \texttt{PAT-SPECTER}, a Sentence Transformers–based model trained using triplet loss on patent citation triples (focal–positive–negative), specifically designed to capture cross-document similarity within the patent domain.

Building on this line of work, \citet{guellec2024npl} present two key findings. First, they show that computing semantic similarity using the full text of patents and scientific articles substantially improves the correspondence with actual citation links, compared to using only titles and abstracts. Second, they compare multiple document embedding models, including \texttt{SPECTER2} and \texttt{PAECTER} (a BERT-for-patents model fine-tuned in the same way as \texttt{PAT-SPECTER}) \citep{ghosh2024paecter}, and find limited performance differences between them when used with a simple CLS-token representation, suggesting that model architecture may be less critical than input richness (e.g., full text vs. title and abstract).

However, given the scale of the data involved in this research and the substantial stylistic and rhetorical differences between patents and other kinds of documents, direct similarity scoring is computationally intensive and may not yield interpretable results \citep{xu2019novel}. 
To address this challenge, we propose a concept-based alignment approach using LLMs. Specifically, we propose an original solution: leveraging the ability of LLMs to extract structured semantic representations from unstructured text, a capability that has been successfully demonstrated across multiple domains including scientific literature \citep{dagdelen2024structured}. Further, recent work has shown that LLMs can outperform traditional sequence labeling models on information extraction tasks when guided by schema-aware or instruction-based prompts \citep{wiest2024llm, han2023empirical}. In the patent context, the ability of LLMs to generate meaningful summaries for downstream tasks has been proven by \citet{yoshikawa2025large}, where the authors create summaries of patent text using GPT-3.5-turbo and see improvements in F1 scores compared to the original text.

Specifically, we adopt an ontology-guided strategy informed by the patent ontology developed by \citet{zhai2022patent}. Using prompt engineering and in-context learning (ICL), we extract three core conceptual categories from both SDG papers and patents titles and abstracts:\footnote{For this extraction task, we concatenate titles and abstracts of both papers and patents in a single text.}

\begin{enumerate}
    \item \textit{Functions}: These represent the primary technological problems or objectives addressed by the invention. Linguistically, they often appear in ``verb + object'' structures, such as ``reduce emissions'' or ``detect anomalies,'' and indicate the intended functional outcome.
    
    \item \textit{Solutions}: These refer to the technical mechanisms or processes proposed to fulfill the identified functions. In text, they typically follow expressions like ``used for,'' ``to achieve,'' or ``by means of,'' describing the means by which the function is realized.
    
    \item \textit{Applications}: These denote the contextual or industrial domains in which the solution is expected to be implemented (e.g., healthcare, transportation, agriculture). Applications often appear toward the end of abstracts or descriptions, introduced by phrases such as ``such as'' or ``e.g.''.
\end{enumerate}

We adopt a patent-centered ontology because these concepts are both explicitly and structurally represented in patent documents and can be meaningfully searched for in other texts. In this sense, our approach also represents an evolution of recent literature that uses NER models to extract concepts from text to compare patents to other documents such scientific publications \citep{hain2022mapping}. 

From a technical standpoint, we design a prompt for the \texttt{gpt-4.1-mini} model \citep{openai2024gpt4technicalreport}, tasking it with extracting the desired elements from both scientific publications and patent texts in the \emph{Patent-SDG Dataset}. The \texttt{gpt-4.1-mini} model is chosen for its state-of-the-art capabilities and cost efficiency.\footnote{\url{https://openai.com/index/gpt-4-1/}}
At the conclusion of this step of the pipeline, we obtain, for each patent, a structured set of extracted \textit{functions}, \textit{solutions}, and \textit{applications}. A corresponding set of concepts is constructed for each SDG by aggregating LLM-extracted elements from the SDG publications. These structured representations form the basis for the concept-level alignment procedure described in the next paragraph.

\subsubsection*{Semantic similarity and ranking}

The second component of our labeling function focuses on associating each patent with one or more SDGs by semantically aligning structured concepts extracted from patent texts and SDG-labeled documents. 

First we create word embeddings for these concepts.\footnote{To encode the concepts we use \texttt{PAECTER} \citep{ghosh2024paecter}, further presented in Section \ref{inter_eval}.} Then, for each patent, we calculate semantic similarity scores $S_i(p, c)$ between its concept(s) and all candidate concepts $c$ from SDG-labeled publications, across each category $i \in \{ \mathrm{function}, \mathrm{solution}, \mathrm{application} \}$.  
Recognizing that the semantic spaces and similarity scales across these categories differ, we introduce three (function-solution-application) \textit{independent semantic thresholds} $\tau_i$, one per category, filtering out low-confidence semantic similarity matches. Then, after the filtering, for each category we create a ranking, sorted in descending order according to the similarity score, that contains the remaining SDG concepts. The reason for not considering the similarity scores as they are is that we expect different similarity ranges across the three categories and therefore, to combine them effectively we use a ranking logic.

We then want to unify the results of the three ranking into a single one. For this purpose, we apply \textit{Reciprocal Rank Fusion (RRF)} to combine the filtered rankings:
\begin{equation}
\mathrm{RRF}(d) = \sum_{i \in \{\mathrm{func},\, \mathrm{sol},\, \mathrm{app}\}} \frac{1}{k + \mathrm{rank}_i(d)}
\end{equation}
where $d$ is a SDG document, $\mathrm{rank}_i(d)$ is its position in the thresholded ranking for category $i$, and $k$ is a dampening constant. 

From the fused ranking, we select the top-$N$ aligned SDG documents for each patent.  
Since each SDG document is pre-annotated with one or more SDGs, we derive the patent’s normalized SDG vector composed by the normalized SDG frequency for each SDG:
\begin{equation}
P(\mathrm{SDG}_j | p) = \frac{\mathrm{count\,of\,SDG}_j\, \mathrm{in\,top}\, N}{\mathrm{top}\,N}
\end{equation}

Therefore, for each patent $p$ we obtain a vector of length 17 where each component of the vector is the frequency of the corresponding SDG in the top-$N$ fused ranking normalized with the value of the hyperparameter top\_$N$ itself.\footnote{For instance, if the top-$N$ selected value is 5, the vector of a patent that has 3 SDG2 and 2 SDG4 in the final fused ranking will be [0, 0.6, 0, 0.4, 0, 0, 0, 0, 0, 0, 0, 0, 0, 0, 0, 0, 0]}
This vector serves as a soft-relevance profile rather than a strict probabilistic prediction, reflecting the relative strength of alignment between the patent and each SDG. 

\subsubsection*{Weakly-supervised optimization}

A central feature of weak supervision is the ability to calibrate noisy, heuristic signals against partial or indirect supervision sources, in order to produce a more reliable training signal \citep{Ratner_2017, geng2016labeldistributionlearning}. In our case, the labeling function \(\lambda(p)\), which assigns a soft 17-dimensional SDG relevance vector to each patent, is parameterized by several interpretable hyperparameters: three semantic similarity thresholds \(\tau_{\text{function}}, \tau_{\text{solution}}, \tau_{\text{application}}\), and a top-$N$ cutoff used to derive SDG distributions from ranked concept alignments.

To tune these parameters, we treat the calibration process itself as a weak supervision task. Specifically, we use 40\% of the \emph{Patent-SDG Dataset} as a validation set for optimization, stratified by SDG distribution. Importantly, this calibration does not attempt to reconstruct the exact citations, but instead evaluates whether the labeling function is able to recover the correct SDG themes based on conceptual alignment. To this end, we consider the SDG labels of cited scientific papers as \textit{positive-only} targets, meaning that we do not assume that uncited SDGs are irrelevant or incorrect, but only that cited ones reflect known, meaningful associations. This is consistent with the idea of \textit{incompleteness} of NPL stressed in the literature \citep{narin1997increasing, he2007evidence}.

To quantify alignment with these positive targets, we define the following loss function:

\begin{equation}
\mathrm{MSE}_{\mathrm{pos}} = \frac{1}{|\mathcal{P}|} \sum_{p \in \mathcal{P}} \frac{1}{|S_p^+|} \sum_{j \in S_p^+} \left( \hat{P}(\mathrm{SDG}_j | p) - P(\mathrm{SDG}_j | p) \right)^2
\end{equation}

Here, \(\mathcal{P}\) denotes the set of validation patents, \(S_p^+\) is the set of SDGs associated with the scientific publications cited in patent \(p\), \(\hat{P}(\mathrm{SDG}_j | p)\) is the normalized SDG distribution predicted by our labeling function, and \(P(\mathrm{SDG}_j | p)\) is the normalized empirical distribution derived from the citations. Crucially, this objective focuses only on the known positive dimensions, reflecting the weak supervision assumption that unlabeled targets may be positive but unobserved.
\footnote{We leverage the \textit{Optuna} Python Package as optimization framework \citep{akiba2019optunanextgenerationhyperparameteroptimization}. In particular, we use Tree-structured Parzen Estimator (TPE), a Bayesian optimization method, widely used in recent parameter tuning frameworks. The results of the optimization process are $\tau_{function}$: 0.260, $\tau_{solution}$: 0.165, $\tau_{application}$: 0.873. The top $N$  optimized threshold is 30. \label{fn:opt}}
Once the hyperparameters are optimized, the labeling function \(\lambda(p)\) is applied to the remaining 60\% of the patents in the Patent-SDG data set. The resulting SDG vectors serve as the silver-standard annotations used to train the downstream supervised models, as described in the following section.

\subsubsection{Mapping patents to SDG as a multi-label regression task}\label{sec:class}

Building upon the silver dataset derived by the LF, we approach the classification of patents into SDGs as a multi-label regression problem. This choice is motivated by the intrinsic characteristics of our generated labeled dataset, where assigned SDG labels are not strictly binary but reflect continuous strengths of association obtained from semantic similarity and rank fusion. The resulting relevance distributions explicitly encode the relative importance of each SDG for a given patent, thereby justifying a regression-based modeling approach rather than traditional binary multi-label classification.
To effectively model these probabilistic outputs, we propose a neural network-based multi-output regression architecture. Specifically, our model comprises a transformer-based embedding layer followed by fully connected layers that independently predict continuous-valued strengths for each of the 17 SDGs. Each output directly corresponds to the normalized relevance distributions derived from the fused SDG rankings described previously. A final ReLU activation function ensures non-negative predictions \citep{agarap2018deep}, consistent with the semantics of probability-like distributions and SDG importance scores. The model takes raw patent text, composed by a concatenation of its title and abstract, as input and predicts a 17-dimensional continuous SDG relevance vector, aligning with the labels generated by our LF.

The neural network is trained using the Mean Squared Error (MSE) loss function, formulated as:
\begin{equation}
\mathrm{MSE} = \frac{1}{N} \sum_{p=1}^{N}\sum_{j=1}^{17}\left(y_{pj} - \hat{y}_{pj}\right)^2
\end{equation}
where \(y_{pj}\) represents the true probabilistic value for SDG \(j\) associated with patent \(p\), and \(\hat{y}_{pj}\) denotes the model’s predicted value. This loss function directly optimizes the model to approximate accurately the continuous strengths of each SDG, respecting the fluid and continuous nature of the labels.
The decision to utilize multi-output regression rather than alternative modeling methods, such as softmax-based probability distributions or sigmoid-based binary classifiers, stems primarily from the necessity to independently preserve the absolute strength of each SDG. Softmax-based approaches would impose artificial mutual exclusivity among labels, forcing a competitive structure that misrepresents the genuine nature of SDG associations. Conversely, the multi-output regression framework naturally aligns with the continuous strengths derived from weighted reciprocal rank fusion, facilitating interpretability by explicitly modeling the independent relevance of each SDG. Moreover, this regression-based approach aligns closely with our dataset generation methodology, as the SDG labels inherently represent continuous strengths rather than categorical assignments. By directly modeling these continuous labels, we gain clearer insights into the degree of relevance between patents and each SDG, enhancing interpretability for stakeholders who seek precise and nuanced assessments of SDG alignment.

\section{Evaluation}\label{sec:valid}

In this Section, we assess the quality of the silver-standard labels generated by our WS pipeline. Since the primary goal of our approach is to build upon the existing SDG citation signal, while acknowledging its limitations, we begin by evaluating how well our LF recovers known SDG associations. To do so, we compare it against several baseline methods, including standard information retrieval (IR) techniques, fine-tuned transformer models, and a zero-shot prompting approach, using \textit{recall} as the key metric. This allows us to test whether the silver signal preserves the original signal (Section \ref{inter_eval}).\\

However, recall-based evaluation only captures the ability of a method to recover existing SDG assignments (i.e., true positives (TP)), and does not account for the correctness of newly generated labels, which could be either true or false positives (FP). Since no  ground-truth exists, this makes direct validation of additional labels particularly challenging. To address this, we implement a validation strategy that uses exogenous patent metadata structured in a network form. By constructing three networks—based on patent citations, co-inventorship, and co-applicancy, among patents in the silver dataset, we evaluate how well different labeling approaches align with the latent structure of each network. We use overlapping modularity as a proxy for label quality. The underlying intuition is that patents connected through these networks are more likely to share thematic, cognitive, or organizational similarities. Therefore, stronger alignment between SDG labels and network homophily suggests higher labeling quality (Section \ref{ext_eval}).

\subsection{Internal evaluation: Comparison with NPL-derived labels}\label{inter_eval}

To assess the quality of the silver-standard labels produced by our LF, we benchmark them against SDG labels derived from NPL, a commonly used reference point in the literature, despite its known limitations \citep{ciarli2022changing}.

We frame this comparison as a multi-label classification task. The goal is to evaluate how well different methods, including our LF and several baseline models, recover the SDGs associated with each patent based on this NPL-derived ground truth.

To evaluate our method, we use the final output of the LF itself: a 17-dimensional SDG relevance vector generated from the full pipeline with optimized hyperparameters. 
We compare our LF to a set of established baseline models, spanning sparse retrieval, transformer-based semantic models, and LLMs. Each baseline is selected to represent a different family of methods for encoding:
\begin{itemize}
    \item \textit{BM25}: BM25 is a well-established probabilistic ranking function used in information retrieval systems, rooted in the Okapi BM25 framework. It computes document relevance based on term frequency, inverse document frequency, and document length normalization. Unlike semantic embedding models, BM25 relies purely on lexical overlap between query and document terms, making it an interpretable and computationally efficient baseline. This method represents bag-of-words methods \citep{robertson1995okapi}.
    \item \textit{BERT-for-Patents}: This model is based on the Bidirectional Encoder Representations from Transformers (BERT) architecture \citep{devlin2019bert}, a deep bidirectional language model that leverages the Transformer encoder to learn contextual representations from large-scale corpora using masked language modeling (MLM) and next sentence prediction (NSP) objectives. The version used here, \texttt{BERT-for-Patents} is pretrained specifically on a large corpus of patent documents, sourced from datasets like USPTO and EPO filings. It represents a pretrained model specific for patent text \citep{srebrovic2020leveraging}.
    \item \textit{SciBERT}: SciBERT is a transformer-based language model built upon the BERT architecture, specifically pretrained on a large corpus of scientific publications drawn from Semantic Scholar. SciBERT retains the standard BERT pretraining objectives, masked language modeling (MLM) and next sentence prediction (NSP), but adapts them to the linguistic and structural characteristics of scientific literature . \texttt{SciBERT} represents a domain-specific pretrained language model for scientific literature \citep{beltagy2019scibert}.
    \item \textit{PAECTER and PAT-SPECTER}: Both PAECTER and PAT-SPECTER are contrastively fine-tuned sentence embedding models trained using triplet loss in a Sentence-BERT (SBERT) framework \citep{reimers2019sentence}. PAECTER builds on \texttt{BERT-for-Patents}, where each training sample consists of an anchor patent, a positively cited patent, and a negative patent that is not cited. In contrast, PAT-SPECTER is based on \texttt{SPECTER2} \citep{singh2022scirepeval}, a transformer model pretrained on scientific articles using citation supervision, and is similarly fine-tuned on examiner citation triplets to adapt it to the patent domain. These models represent domain-specific and citation-informed sentence encoders \citep{ghosh2024paecter}.
\end{itemize}

For all baseline methods, we simulate in fact an equivalent relevance vector using a retrieval-based approach: we concatenate each patent’s title and abstract, encode it using the method in question, and retrieve the top-$k$ most similar SDG-tagged scientific publications, where $k = 30$, matching the threshold used in our LF.\footnote{The threshold of 30 refers to the optimized threshold of the top\_N hyperparameter as discussed in Footnote \ref{fn:opt}.} The SDG vector for each patent is then constructed from the SDG distribution of the retrieved documents.
This setup allows for a fair, standardized evaluation of all methods against the same weak gold standard, enabling a direct comparison of performance in recovering known SDG associations from NPL citations.

In addition to embedding-based retrieval methods, we further evaluate a zero-shot classification baseline using  \texttt{GPT4}.\footnote{We implemented a zero-shot SDG classification baseline using the \texttt{gpt-4.1-mini} model via the \texttt{LangChain} framework. A structured prompt was designed to present the patent's title and abstract to the model, instructing it to assign relevant SDGs from 1 to 17 as a multi-label classification task. SDG definitions were provided as field-level descriptions in a structured schema, which the model used to return outputs via OpenAI's function-calling interface. The model was queried with \texttt{temperature=0} to ensure near-deterministic behavior, and SDGs were marked \texttt{True} only if the invention demonstrated a clear and substantial technical connection to the goal, as per conservative classification guidelines embedded in the prompt.} This baseline serves two purposes: first, it offers a reference point for evaluating the performance of recent LLMs on SDG-related classification tasks in the absence of supervision or retrieval; and second, it provides a diagnostic tool to assess the added value of our weakly supervised LF compared to a purely prompt-based approach. While the zero-shot LLM does not produce document embeddings and is therefore not directly comparable to the encoding-based models, its inclusion highlights the current limitations of general-purpose models in capturing domain-specific relevance signals for complex, multi-label classification tasks.

Because we aim to assess whether the silver signal captures known SDG associations, we focus on \textit{recall} as the primary metric. Unlike precision or F1, recall does not penalize predictions beyond the citation-based labels, which is important in this context. Since NPL-derived SDGs are incomplete and cannot be treated as ground-truth negatives, we cannot confidently identify false positives. Therefore, evaluating precision or computing F1 would underestimate the true quality of predictions.

Recall is computed as follows:
\begin{equation}
\text{Recall} = \frac{\text{True Positives}}{\text{True Positives} + \text{False Negatives}}
\end{equation}
where true positives are SDG labels correctly recovered by the model (already present in the NPL-based set), and false negatives are labels present in the NPL reference but missed by the models.

\begin{table}[htbp]
\centering
\caption{Recall performance on SDG prediction using NPL citation signal as reference}
\label{tab:recall_only}
\begin{tabular}{lcc}
\toprule
\textbf{Model} & \textbf{Macro Recall} & \textbf{Micro Recall} \\
\midrule
PAECTER       & 0.705 & 0.900 \\
PAT-SPECTER    & 0.706 & 0.900 \\
SciBERT       & 0.581 & 0.867 \\
BERT-for-Patents  & 0.530 & 0.843 \\
BM25          & 0.658 & 0.886 \\
GPT-4.1 (LLM) & 0.293 & 0.650 \\
\rowcolor{gray!15}
Silver & \textbf{0.711} & \textbf{0.902} \\
\bottomrule
\end{tabular}

\scriptsize
\emph{Notes:} Recall is computed using binarized vectors. A predicted SDG is counted as correct if it appears in the top 30 output and is present in the NPL-based reference. Macro recall is unweighted; micro recall is weighted by SDG frequency.
\end{table}

The silver-standard produced by the LF achieves the highest performance across both recall metrics, outperforming all baselines. With a macro recall of 0.711 and a micro recall of 0.902, it demonstrates the strongest alignment with the NPL-derived SDG signal, indicating its capacity to recover known SDG associations across both high- and low-frequency classes. Notably, it outperforms fine-tuned transformer models such as \texttt{PAECTER} and \texttt{PAT-SPECTER}, as well as the BM25 bag-of-words baseline.

General-purpose models pretrained on scientific corpora, such as \texttt{SciBERT} and \texttt{BERT-for-Patents}, exhibit lower recall, particularly on macro metrics, suggesting reduced sensitivity to infrequent SDGs. The zero-shot \texttt{GPT-4.1} model yields the weakest performance, reinforcing existing findings that prompt-only approaches are limited in handling complex, domain-specific, multi-label classification tasks without task-specific adaptation \citep{moradi2021gpt, mu2023navigating}.

The comparison with \texttt{PAECTER} is particularly instructive: while \texttt{PAECTER} performs competitively on micro recall (0.900), our LF surpasses it on both macro and micro recall. This suggests that structured semantic abstraction, via concept extraction and retrieval over curated representations, improves generalization across diverse SDG classes, including less common ones. By aligning patents and SDG-tagged publications in a shared conceptual space, the LF enhances comparability beyond what is achievable using raw text embeddings alone.
To illustrate the alignment in practice, Table~\ref{tab:example_sdg6} presents a representative example of conceptual alignment between a patent and a scientific article both linked to SDG 6 (Clean Water and Sanitation). The pair includes the patent US10689267B2, titled “Method for removing organic dyes from water using a hemoglobin/Fe$_3$O$_4$ composite adsorbent”, and the scientific publication “Adsorption of Methylene Blue, Bromophenol Blue, and Coomassie Brilliant Blue by $\alpha$-chitin nanoparticles” by \citet{dhananasekaran2016adsorption}.

Despite differences in formulation and terminology, the LLM-extracted representations highlight consistent underlying concepts: removal of dyes, adsorptive mechanisms, and wastewater treatment. While the materials differ (chitin vs. hemoglobin/Fe$_3$O$_4$), the function and application are clearly aligned. This example illustrates how the LLM-based approach captures conceptual similarity across structurally and stylistically divergent documents, supporting the model's utility for cross-domain semantic alignment more than using raw text.

\begin{table}[ht]
\centering
\caption{Conceptual alignment example between an SDG-tagged scientific publication and a cited patent (SDG 6)}
\label{tab:example_sdg6}
\bigskip
\renewcommand{\arraystretch}{1.4}
\begin{tabular}{|p{3cm}|p{5.5cm}|p{5.5cm}|}
\hline
\rowcolor{gray!20}
\textbf{Extracted \newline element} & \textbf{SDG Publication} & \textbf{Cited Patent} \\
\hline
\textbf{Function} & Remove dyes from effluents & Remove organic dyes and contaminants from water \\
\hline
\textbf{Solution} & Use $\alpha$-chitin nanoparticles derived from shell waste to adsorb dyes such as Methylene Blue, Bromophenol Blue, and Coomassie Brilliant Blue & Use a hemoglobin/Fe$_3$O$_4$ composite adsorbent to bind contaminants, followed by magnetic separation \\
\hline
\textbf{Application} & Wastewater treatment for environmental pollution prevention & Water treatment for removing organic and inorganic contaminants \\
\hline
\end{tabular}
\scriptsize
\newline
\bigskip
\justifying{\emph{Notes:}} The Table presents an example of extracted \emph{function}, \emph{solution} and \emph{application} from patent US10689267-B2 entitled "Method for removing organic dyes from water using a hemoglobin/FE3O4 composite adsorbent
" and the paper entitled "Adsorption of Methylene Blue, Bromophenol Blue, and Coomassie Brilliant Blue by $\alpha$-chitin nanoparticles" by \citet{dhananasekaran2016adsorption}. The paper has been tagged as belonging to SDG6 by the Elsevier SDG Mapping Initiative.
\end{table}

\subsubsection*{Ablation analysis}

To better understand the contribution of each component in our LF, we also conduct an ablation analysis in which we selectively remove or isolate the three semantic concept types used for SDG alignment: \emph{functions}, \emph{solutions}, and \emph{applications}. As described in Section~\ref{sec:abstraction}, our labeling function relies on extracting these three types of concepts from both patents and SDG-labeled scientific documents, and computes semantic similarity scores between them to build a fused SDG relevance signal. This allows us to isolate the effect of using only functions, solutions, or applications (or combinations thereof) on recall performance against the NPL-derived SDG labels.

We evaluate the performance of six alternative versions of the labeling function:

\begin{enumerate}
    \item \textit{Solution \& Application}: removes the \textit{function} component and computes SDG similarity only on solutions and applications.
    \item \textit{Function \& Application}: removes the \textit{solution} component, retaining only functions and applications.
    \item \textit{Solution \& Function}: removes the \textit{application} component, retaining only functions and solutions.
    \item \textit{Functions only}: uses only function-based similarity for SDG scoring.
    \item \textit{Solutions only}: uses only solution-based similarity.
    \item \textit{Applications only}: uses only application-based similarity.
\end{enumerate}

The results are summarized in Table~\ref{tab:ablation}. The full model, labeled \texttt{tot}, represents the complete labeling function that includes all three semantic concept types.

\begin{table}[htbp]
\centering
\caption{Ablation results on SDG prediction (recall only)}
\label{tab:ablation}
\begin{tabular}{lcc}
\toprule
\textbf{Model} & \textbf{Macro Recall} & \textbf{Micro Recall} \\
\midrule
Full (tot)           & 0.711 & 0.902 \\
Solution \& Application          & 0.711 & 0.904 \\
Function \& Application           & 0.691 & 0.894 \\
\rowcolor{gray!15}
Solution \& Function        & \textbf{0.725} & \textbf{0.910} \\
Functions only       & 0.701 & 0.899 \\
Solutions only       & 0.701 & 0.902 \\
Applications only    & 0.653 & 0.880 \\
\bottomrule
\end{tabular}

\scriptsize
\emph{Notes:} Recall is computed using binarized SDG vectors. A predicted SDG is counted as correct if it appears in the top-30 predictions and is also present in the NPL-based reference set.
\end{table}

Removing any single component results in a performance drop or marginal change, confirming that each contributes meaningfully to the quality of the final signal. The largest performance degradation occurs when the \textit{solution} dimension is removed, reducing macro recall to 0.691 and micro recall to 0.894. This suggests that solution-level concepts capture core technological mechanisms relevant for SDG alignment.
When tested in isolation, all three components perform worse than the full model. The \textit{applications only} variant performs the weakest (macro recall 0.653), while \textit{functions only} and \textit{solutions only} both achieve macro recall above 0.70. This confirms that while each concept type is informative on its own, their combination is necessary for optimal performance.
Overall, the ablation results validate the design of the labeling function: combining multiple semantic dimensions leads to more comprehensive and robust SDG retrieval from patent texts.
However, removing the \textit{application} component slightly improves recall (macro 0.725, micro 0.910), indicating that this dimension may introduce more noise than signal in some cases, possibly due to its broader or less specific nature. Therefore, we remove this dimension in the final construction of the silver dataset.

\subsection{External evaluation: Evaluation of novel SDG signal through \textit{network homophily}} \label{ext_eval}

To go beyond recall, as recall alone cannot distinguish whether newly identified positives are genuine or false positives, we turn to the idea \textit{network homophily} to evaluate labeling quality in the absence of a ground-truth. Network homophily is the tendency of nodes in a network to form connections with other nodes that are similar to themselves in some relevant attribute \citep{mcpherson2001birds}. While on social networks, this may reflect shared interests or demographics, in technological ones, it may reflects thematic or functional similarity \citep{hwang2023evolution}.

Inspired by the work of \citet{bergeaud2017classifying}, who validated semantic versus CPC-based classifications through network modularity, we extend their logic to assess our SDG labeling over multiple network dimensions including not only citations, but also inventors and applicants. 
When applied to patent networks, the \textit{homophily} idea implies that patents that are closely linked either by citations, shared inventors, or common applicants, are more likely to belong to the same or similar categories. In particular, the idea of  \textit{citation homophily} has been already empirically validated in multiple studies \citep{ciotti2016homophily, nomaler2019greentech}. Furthermore, the intuition behind network homophily can be extended beyond citations to other bibliometric information of patents, such as inventors and applicants \citep{arts2018text,yoo2023novel}. In this sense, patents that have common inventors are likely to be technologically related, as the inventive capacity of individuals is typically concentrated within a limited range of technological domains. Inventors rarely operate across completely unrelated fields, so co-inventorship  may be interpreted as  signal of technological and cognitive proximity \citep{whalen2020patent, fritz2023modelling}. Likewise, patents associated with the same applicant  may reflect the the technological specialization of that organization, as patent portfolios that show a degree of thematic coherence have been proven to hold greater strategic value \citep{parchomovsky2005patent}.

Building on these three relational dimensions, we can assess the quality of different SDG labeling schemes by examining how well their assigned categories align with the community structure of patent networks. If labels are semantically meaningful, then patents connected through citations, inventors, or applicants should exhibit a higher probability of sharing similar SDG distributions. Conversely, if a labeling scheme produces weak or noisy assignments, the induced communities will misalign with the observed network structure, leading to lower homophily.

To operationalize this idea, we rely on \emph{overlapping modularity}, an extension of the standard modularity measure that accommodates nodes belonging to multiple communities. Traditional modularity assumes disjoint partitions, but our setting requires a more flexible approach, as patents may contribute to multiple SDGs with varying degrees of relevance. We adopt the \emph{overlapping modularity} formulation of \citet{nicosia2009extending}, which allows each node to participate in multiple communities with weighted membership, consistently with the continuous relevance scores produced by our LF.

In the work of \citet{nicosia2009extending}, let $G = (V, E)$ denote a  graph, where $V$ is the set of nodes and $E \subseteq V \times V$ is the set of citation links (excluding self-loops and duplicate edges). To represent overlapping community structure, an array of \emph{belonging factors}: 
\[
[\alpha_{i,1}, \alpha_{i,2}, \ldots, \alpha_{i,|C|}]
\]
is assigned to each node $i \in V$, where each coefficient $\alpha_{i,c}$ expresses the strength of node $i$'s association with community $c$. These belonging factors allow a node to participate in multiple communities with varying degrees of membership, rather than being restricted to a single community.

For each community $c$, we compute its contribution to the total modularity using the formula:
\[
Q_c = \frac{1}{m} \sum_{(i \to j) \in E} f(w_{ic}, w_{jc}) 
    - \frac{1}{m^2} \left( \sum_i k_i^{\text{out}} \, \beta_i^c \right) \left( \sum_j k_j^{\text{in}} \, \beta_j^c \right),
\]
where:
\begin{itemize}
    \item $m = |E|$ is the total number of directed edges
    \item $k_i^{\text{out}}$ and $k_j^{\text{in}}$ are the out-degree and in-degree of nodes $i$ and $j$, respectively
    \item $f(w_{ic}, w_{jc})$ is a similarity function applied to the community weights of nodes $i$ and $j$\footnote{In our implementation, we use the \textit{power function} for computing node affinities:
\[
f(w_{ic}, w_{jc}) = (w_{ic})^p \cdot (w_{jc})^p,
\]
where the exponent $p \in (0, 1]$ adjusts the emphasis on partial memberships. Smaller values of $p$ emphasize nodes with weaker memberships. We set \( p = 0.35 \) in the power-based similarity function \( f(a, b) = a^p \cdot b^p \) to account for the unbalanced distribution of SDG labels in our dataset. Lower values of \( p \) amplify the influence of small membership weights, thereby ensuring that rare or weakly expressed SDGs are not disproportionately downweighted in the modularity computation. This is particularly important in our setting, where some SDGs are sparsely represented but may still capture meaningful thematic structure. By using \( p = 0.35 \), we strike a balance between sensitivity to partial memberships and maintaining the robustness of the community signal, while avoiding the tendency of higher \( p \) values to penalize rare contributions in respect of modularity. \label{fn:power}
}
    \item $\beta_i^c$ and $\beta_j^c$ measure the average affinity of nodes $i$ and $j$ to the rest of the network within community $c$\footnote{The $\beta$ terms are defined as:
\[
\beta_i^c = \frac{1}{|V|} \sum_{u \in V} f(w_{ic}, w_{uc}),
\quad
\beta_j^c = \frac{1}{|V|} \sum_{u \in V} f(w_{uc}, w_{jc}).
\]
}
\end{itemize}

The total modularity is computed by summing over all communities:
\[
Q = \sum_c Q_c.
\]

This formulation captures the extent to which the network topology aligns with the overlapping community labels. Labeling schemes that better reflect the underlying structure of the network are expected to produce higher modularity values.

To implement this approach, we construct three distinct  networks using the patents in our silver standard set (\(N=32{,}895\) nodes). In each case, patents are represented as nodes, and edges are introduced according to the relational dimension of interest: citations, inventors, or applicants.  

\begin{itemize}
    \item \textit{Citation network}: An edge is created between two patents if one cites the other.\footnote{We only consider applicant citations in our analysis, because all of the top-performing models presented in Section \ref{inter_eval} and evaluated in this Section are trained specifically on examiners citations, as explained in \citet{ghosh2024paecter}. Therefore, to avoid any overlap with training data and ensure the integrity of our evaluation, we focus exclusively on applicant-provided citations, which are also considered by the literature to be more thematically relevant than those added by examiners. \citep{jaffe1993geographic, criscuolo2008does}
} Within our silver dataset, we observe \(68{,}102\) citation edges, corresponding to \(7,251\) patents that cite at least one other patent in the corpus. This network captures direct technological linkages via knowledge flows.  

    \item \textit{Inventor network}: An edge is created between two patents if they share at least one inventor. This results in \(64{,}463\) inventor edges, with \(12{,}643\) patents having at least a shared connection with another. This indirect network highlights collaborative and cognitive proximity, reflecting the limited technological scope of individual inventors.  

    \item \textit{Applicant network}: An edge is created between two patents if they share at least one applicant. This indirected network is substantially denser, with \(998{,}322\) edges, where \(25{,}391\) patents share at least one applicant with another.  
\end{itemize}

\begin{figure*}[t]
  \centering
  \begin{subfigure}{0.33\textwidth}
    \centering
    \includegraphics[width=\linewidth]{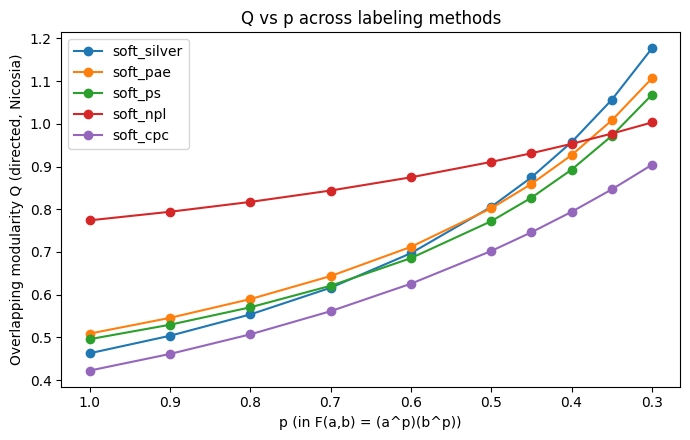}
    \caption{}\label{fig:qvsp-citation}
  \end{subfigure}\hfill
  \begin{subfigure}{0.33\textwidth}
    \centering
    \includegraphics[width=\linewidth]{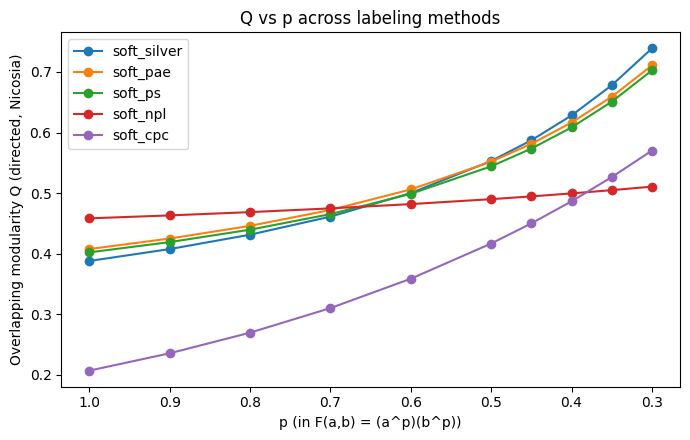}
    \caption{}\label{fig:qvsp-applicant}
  \end{subfigure}\hfill
  \begin{subfigure}{0.33\textwidth}
    \centering
    \includegraphics[width=\linewidth]{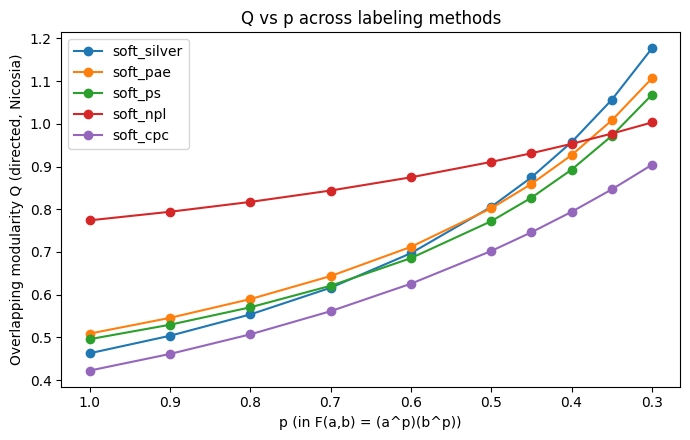}
    \caption{}\label{fig:qvsp-cpc}
  \end{subfigure}

  \caption{Overlapping modularity \(Q(p)\) (directed Nicosia; \(F(a,b)=(a^p)(b^p)\)) across labeling methods.
  Panels: \subref{fig:qvsp-citation} citation graph, \subref{fig:qvsp-applicant} inventor graph, \subref{fig:qvsp-cpc} applicant graph.}
  \label{fig:qvsp-all}
\end{figure*}

By analyzing \textit{modularity} across these three networks, we are able to evaluate the extent to which labeling schemes align with different forms of structural homophily: \emph{thematic proximity} (citations), \emph{cognitive proximity} (inventors), and \emph{organizational proximity} (applicants). A labeling scheme that consistently achieves higher  modularity across all three networks can be interpreted as providing semantically coherent and structurally validated SDG assignments. 

\begin{table}[htbp]
\centering
\caption{Overlapping modularity scores across citation, inventor, and applicant networks for different labeling schemes}
\label{tab:modularity_networks}
\begin{tabular}{lccc}
\toprule
\textbf{Labeling Scheme} & \textbf{Thematic proximity} & \textbf{Cognitive proximity} & \textbf{Organizational proximity} \\
\midrule
CPC           & $0.84$ & $0.92$   & $0.52$ \\
NPL           & $0.97$ & $0.85$ & $0.50$ \\
\rowcolor{gray!15}
Silver        & $\mathbf{1.05}$ & $\mathbf{0.98}$ & $\mathbf{0.68}$ \\
PAECTER       & $1.00$   & $0.95$   & $0.66$ \\
PAT-SPECTER    & $0.97$   & $0.94$   & $0.65$ \\
\bottomrule
\end{tabular}
\scriptsize
\justifying
\emph{Notes:} Modularity values are reported for three relational dimensions of the patent system. 
\textit{Thematic proximity} refers to citation links, capturing technological relatedness through knowledge flows. 
\textit{Cognitive proximity} is measured via inventor co-assignments, reflecting the bounded expertise of individual inventors. 
\textit{Organizational proximity} is derived from shared applicants, representing the technological specialization of firms. 
Higher modularity indicates stronger alignment between SDG labels and the underlying network structure. 
The presented results refers to \( p = 0.35 \), as previously discussed in Footnote \ref{fn:power}.
\end{table}

Table \ref{tab:modularity_networks} compares the modularity scores of five labeling schemes across the three network dimensions. CPC and NPL serve as baselines: CPC represents the standard technological taxonomy, while NPL corresponds to the weak supervision source. We contrast these with the \texttt{Silver} signal and the two best models from the internal evlauation, \texttt{PAECTER} and \texttt{PAT-SPECTER}.

Three patterns emerge from the results.

First, the \texttt{Silver} labels achieve consistently high modularity across all networks, with values of 1.05 for citations and 0.98 for inventors, and 0.68 for applicants. This indicates that the silver standard assignments are the most coherent across thematic, cognitive, and organizational dimensions.

Second, out-of-the-box similarity methods (\texttt{PAECTER} and \texttt{PAT-SPECTER}) also show strong performance. In particular, \texttt{PAECTER} reaches a high score in the citation network (1.00) and performs competitively in inventor (0.95) and applicant (0.66) networks. These results suggest that transformer models capture meaningful semantic regularities, though they still fall short of the structural coherence achieved by \texttt{Silver}, especially on the applicant dimension.

Third, CPC and NPL provide weaker baselines. While CPC achieves relatively strong performance in the inventor dimension (0.92), its citation (0.84) and applicant (0.52) scores remain lower. NPL fares similarly, with slightly higher citation modularity (0.97) but the weakest applicant alignment (0.50). This confirms that while useful for broad categorization or weak supervision, neither scheme provides sufficiently fine-grained or thematically coherent SDG signals.

Furthermore, Figure~\ref{fig:qvsp-all} reports the values of overlapping modularity \(Q(p)\) as a
function of the power parameter \(p\) for three graphs (citations, inventors, applicants). We compare the \texttt{Silver} labels against the baselines. The general shape of the curves is consistent across all graphs:
as \(p\) decreases, all methods show a monotone increase in modularity, reflecting the
fact that lowering \(p\) amplifies the contribution of weak overlaps.\footnote{For \(0<p<1\), small values of \(\alpha_{i,c},\alpha_{j,c}\) are magnified by the
transformation \(F_p(\alpha_{i,c},\alpha_{j,c})=(\alpha_{i,c}^p)(\alpha_{j,c}^p)\).
The observed term in \(Q(p)\) sums \(F_p(\alpha_{i,c},\alpha_{j,c})\) over actual edges,
so edges with weak but shared memberships contribute much more as \(p\) decreases.
The expected term also increases, but only through the average prevalence
\(\overline{\alpha_c^p}=\tfrac{1}{|V|}\sum_u \alpha_{u,c}^p\), which is diluted across all nodes.
Hence, when weak memberships are concentrated on edges, the observed growth exceeds
the null, and \(Q(p)\) rises.}Two regimes are apparent. Near \(p=1\), where only strong SDG memberships matter,
\texttt{npl} and the baselines outperform \texttt{silver}, indicating that
they capture the dominant SDG signal that explains a large share of links. As \(p\)
decreases into the range \(0.5\)–\(0.35\), the slope of \texttt{Silver} becomes
steeper and the method consistently overtakes all baselines. This effect is most
pronounced on the citation graph, but is also visible on the applicant and inventor graphs. The interpretation is that the low--probability, secondary
SDG assignments produced by \texttt{Silver} are not random noise but align with the
observed edges more than expected under the null. In contrast, the tails of the
baselines and the \texttt{NPL} method do not yield the same advantage. Finally, the
\texttt{CPC} taxonomy rises with \(p\) but remains well below the SDG--based methods,
indicating that CPC may capture a different dimension of relatedness.

Overall, the stability of this pattern across three different graphs confirms that our
method is not only competitive in capturing primary SDG signals, but also adds value
in the weak--signal regime, where secondary SDGs provide genuine structural
information about the connectivity of the patent network.

Taken together, these results demonstrate that the \texttt{Silver} approach yields the most structurally consistent labeling, outperforming both traditional classifications and state-of-the-art transformer-based methods across the multiple dimensions of patent network homophily.

\clearpage
\section{Results}\label{sec:res}

This Section presents results of the research. First, we present the silver standard dataset with some stylized facts (Section \ref{sec:silvsta}).
We then evaluate several models trained on the silver-labeled data, leveraging different transformer-based architectures (Section \ref{sec:down}). In particular, we assess whether the labeling function produces a signal that is learnable across model classes; second, we compare different models performances and identify which architectures are best suited for the patent-to-SDG mapping task.

\subsection{The \textit{silver standard dataset}} \label{sec:silvsta}

The final output of the labeling function is the \textit{silver standard dataset}, composed of 32,895 patents\footnote{This subset does not correspond exactly to 60\% of the original data because the split was performed using stratified sampling to preserve the distribution of SDG labels.}, to which are assigned 150,946 SDG labels, an increase of  more than 240\% if considering the NPL citations assigned only 43,956 SDG labels to the same set of patents. In particular, the avergage number of SDG labels assigned to each patent rises from 1.34 (NPL signal) to 4.59 (silver signal).
\begin{figure}[ht]
\centering

\begin{subfigure}[b]{0.48\textwidth}
    \centering
    \includegraphics[width=\textwidth]{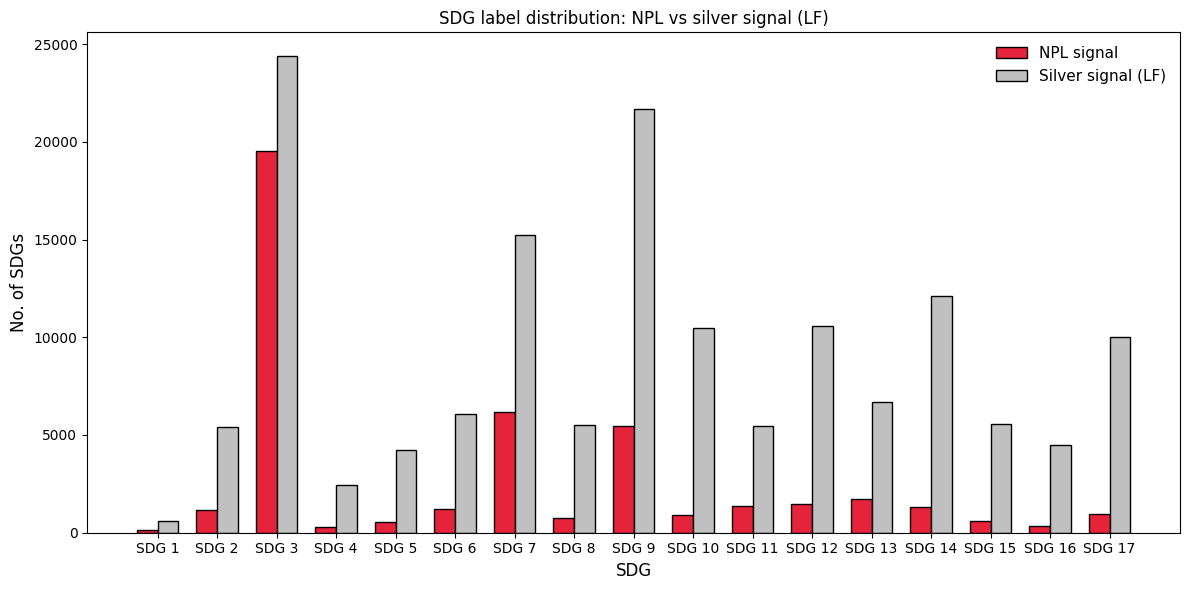}
    \caption{Absolute number of SDG label assignments}
    \label{fig:sdg-distribution-abs}
\end{subfigure}
\hfill
\begin{subfigure}[b]{0.48\textwidth}
    \centering
    \includegraphics[width=\textwidth]{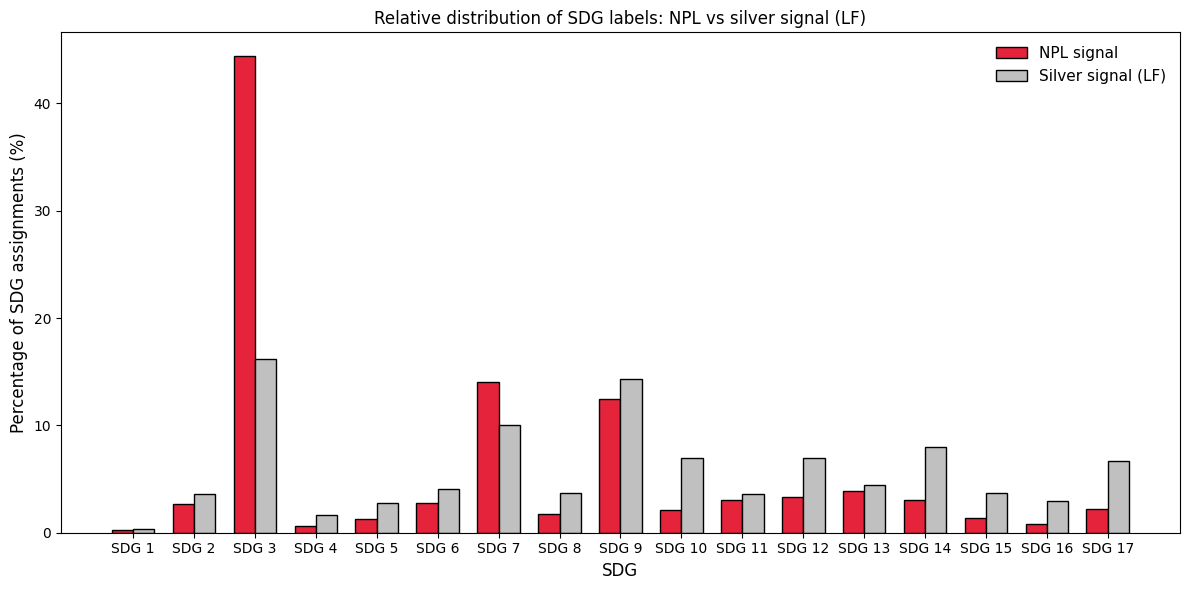}
    \caption{Relative proportion of SDG assignments}
    \label{fig:sdg-distribution-rel}
\end{subfigure}

\caption{Comparison of SDG label distributions between the NPL signal (red) and the silver signal generated by the labeling function (gray). Each bar represents the number of patents (a) or percentage of total assignments (b) associated with each of the 17 SDGs.}
\label{fig:sdg-distribution-comparison}
\end{figure}

Figure~\ref{fig:sdg-distribution-comparison} compares the SDG label distributions obtained via the original NPL citation signal and the silver signal generated by our labeling function. Panel (a) reports the absolute number of SDG assignments, while panel (b) presents the normalized proportions, showing each SDG's relative prevalence.

Overall, the silver signal not only expands the original coverage but also covers a broader and more balanced range of SDGs compared to the highly concentrated distribution of the NPL signal. In particular, SDG 3 (Good Health and Well-being) dominates the NPL signal, accounting for more than 40\% of all SDG assignments, whereas its weight is notably reduced in the labeling ot the silver signal. The labeling function instead distributes relevance across multiple goals, substantially increasing representation for cross-cutting SDGs such as  SDG 10 (Reduced Inequalities), SDG 14 (Life Below Water), and SDG 16 (Peace, Justice and Strong Institutions). Further, we do not have cases where the labeling function fails entirely; that is, no patent in the silver standard dataset receives an all-zero relevance vector.

This distribution may suggests that the silver signal captures broader sustainability dimensions that are underrepresented in the NPL-derived annotations, thus enabling a more comprehensive and diversified SDG mapping across the patent landscape.

\subsubsection*{Alignment between NPL signal and \textit{silver standard dataset}}

Section \ref{inter_eval} already investigated the recall of the LF as compared to the original NPL signal. To assess the similarity of the two SDG distributions more globally, we compute Pearson and Spearman correlation coefficients over the aggregated SDG frequencies. The results show a strong linear correlation (Pearson \(r = 0.81\), \(p < 0.001\)) and high rank-order agreement (Spearman \(\rho = 0.85\), \(p < 0.001\)).

Another important dimension of alignment between the NPL-derived and silver standard labels lies in the structure of SDG co-occurrences. While  frequency alignment reflects agreement in the overall prevalence of individual SDGs, the co-occurrence analysis captures more nuanced relational patterns, specifically, how frequently different SDGs appear together within the same patent. This is particularly relevant given the multidimensional nature of sustainability-related innovation, where technological solutions often contribute simultaneously to multiple SDGs.
To assess this, we compute pairwise SDG-SDG co-occurrence matrices for both the NPL and silver signals. Each cell \((i,j)\) in these matrices represents the number of patents in which both SDG~\(i\) and SDG~\(j\) are present, excluding the diagonal. Since these raw counts are heavily influenced by the overall frequency of each SDG, we normalize the matrices row-wise to obtain conditional co-occurrence profiles. That is, each row of the normalized matrix for SDG~\(i\) is interpreted as a probability distribution over the other SDGs \(j \ne i\), reflecting the relative frequency with which SDG~\(i\) co-occurs with each of the others.

\begin{figure}[htbp]
\centering
\includegraphics[width=0.6\textwidth]{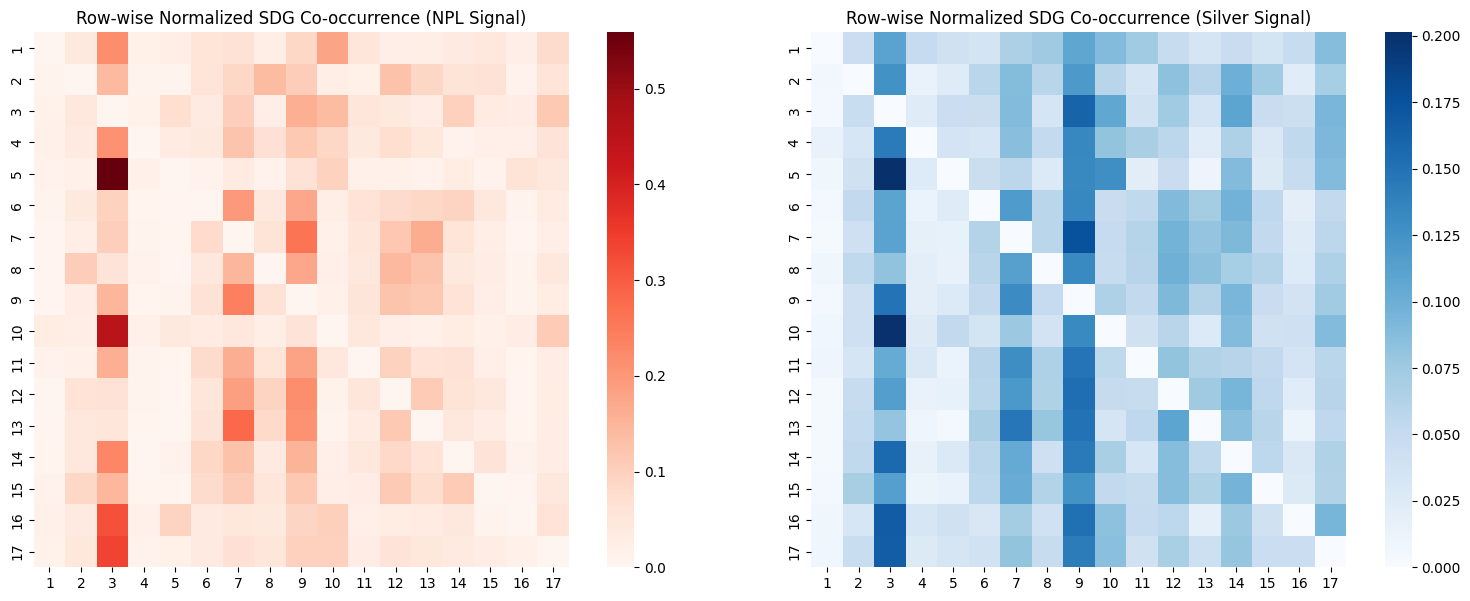}
\caption{Row-wise normalized SDG co-occurrence matrices for the NPL signal (left) and the silver signal (right). Each cell \((i,j)\) represents the conditional probability \(P(j \mid i)\) that SDG~\(j\) appears in a patent given the presence of SDG~\(i\).}
\label{fig:cooccurrence_heatmaps}
\end{figure}

Figure~\ref{fig:cooccurrence_heatmaps} displays the row-wise normalized co-occurrence matrices (with no diagolals) for both the NPL and silver signals.\footnote{The normalization is due to the fact that some SDGs occurr more often than others and this may influence the co-occurrences patterns too.} The comparison of row-wise normalized co-occurrence profiles reveals important thematic consistencies and divergences between the NPL and silver signals. In both matrices, SDG~3 (\textit{Good Health and Well-being}) emerges as a central hub, frequently co-occurring with a broad range of goals including SDG~9 (\textit{Industry, Innovation and Infrastructure}) and SDG~10 (\textit{Reduced Inequalities}). This aligns with the role of health-related technologies in addressing systemic societal and industrial challenges.

SDG~7 (\textit{Affordable and Clean Energy}) also shows strong and consistent co-occurrence with SDG~13 (\textit{Climate Action}) in both signals, reflecting the well-established link between clean energy solutions and environmental mitigation. The silver matrix reinforces this connection more uniformly across other goals as well, such as SDG~12 (\textit{Responsible Consumption and Production}) and SDG~11 (\textit{Sustainable Cities and Communities}), pointing to its ability to capture semantic consistencies beyond the original citation signal. In this sense, also the co-occurrence between SDG~11 (\textit{Sustainable Cities and Communities}) and SDG~13 (\textit{Climate Action}) is markedly stronger in the silver matrix, as well as the co-occurrence between
SDG~14 (\textit{Life Below Water}) and SDG~15 (\textit{Life on Land}).
Taken together, these patterns suggest that the silver labeling method is not only structurally aligned with the NPL signal but may in fact unveil other semantically justified connections among SDGs.
Moreover, to quantify the structural similarity between the two matrices, we compute both Pearson and Spearman correlation coefficients over the  conditional probability distributions resulting from row normalization. The Pearson correlation of \(r = 0.788\) (\(p < 0.001\)) and Spearman correlation of \(\rho = 0.785\) (\(p < 0.001\)) confirm a strong alignment in both the magnitude and rank-order of co-occurrence profiles. These results indicate that, even after controlling for marginal SDG frequencies, the silver labeling function preserves the relational structure of SDG associations as implied by the NPL signal.

\subsubsection*{Comparison with CPC taxonomy}

\begin{figure}[h]
    \centering
    \begin{subfigure}[b]{0.48\textwidth}
        \centering
        \includegraphics[width=\textwidth]{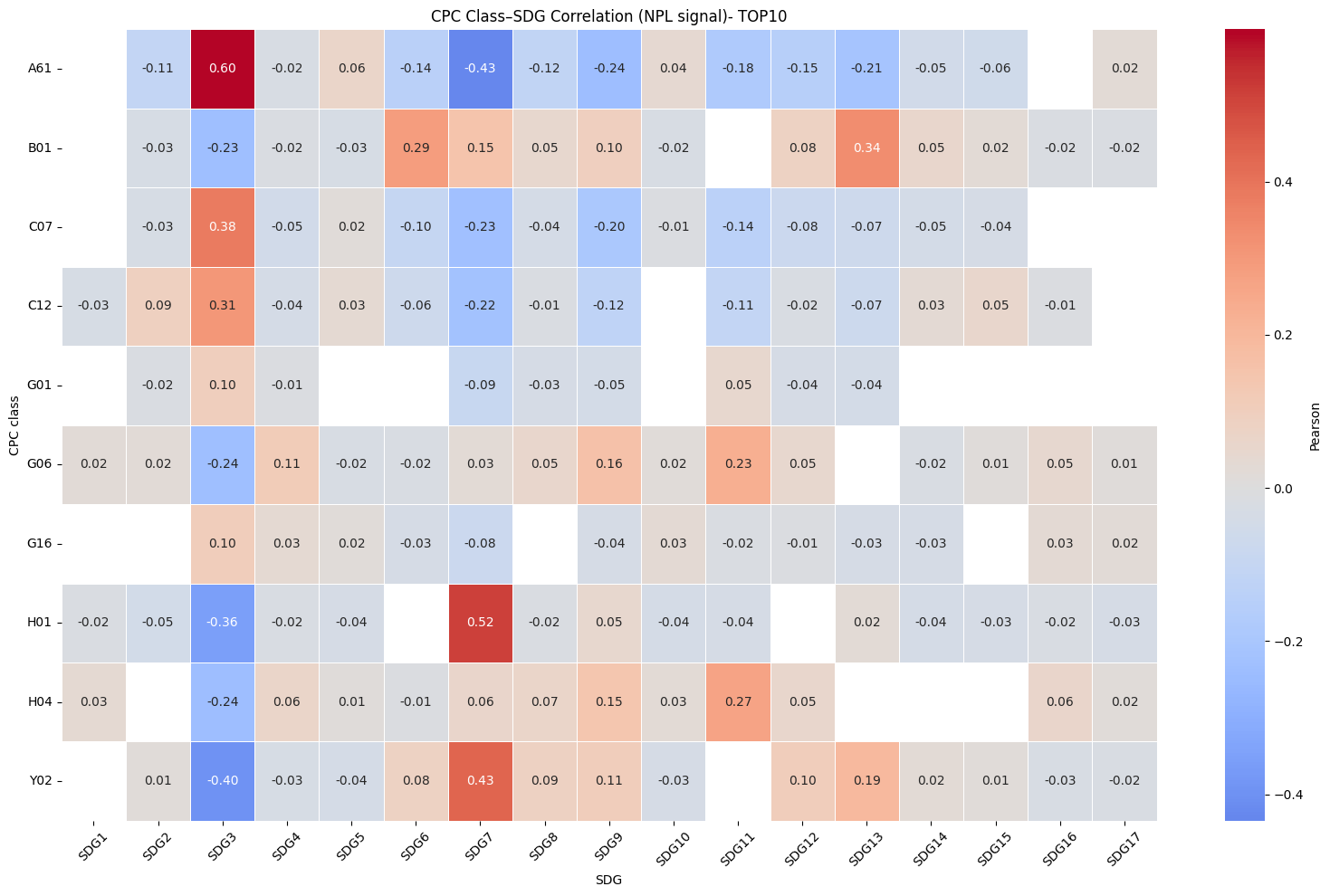}
        \caption{Pearson correlation between CPC classes and SDGs (NPL-based signal)}
        \label{fig:cpc_sdg_corr_npl}
    \end{subfigure}
    \hfill
    \begin{subfigure}[b]{0.48\textwidth}
        \centering
        \includegraphics[width=\textwidth]{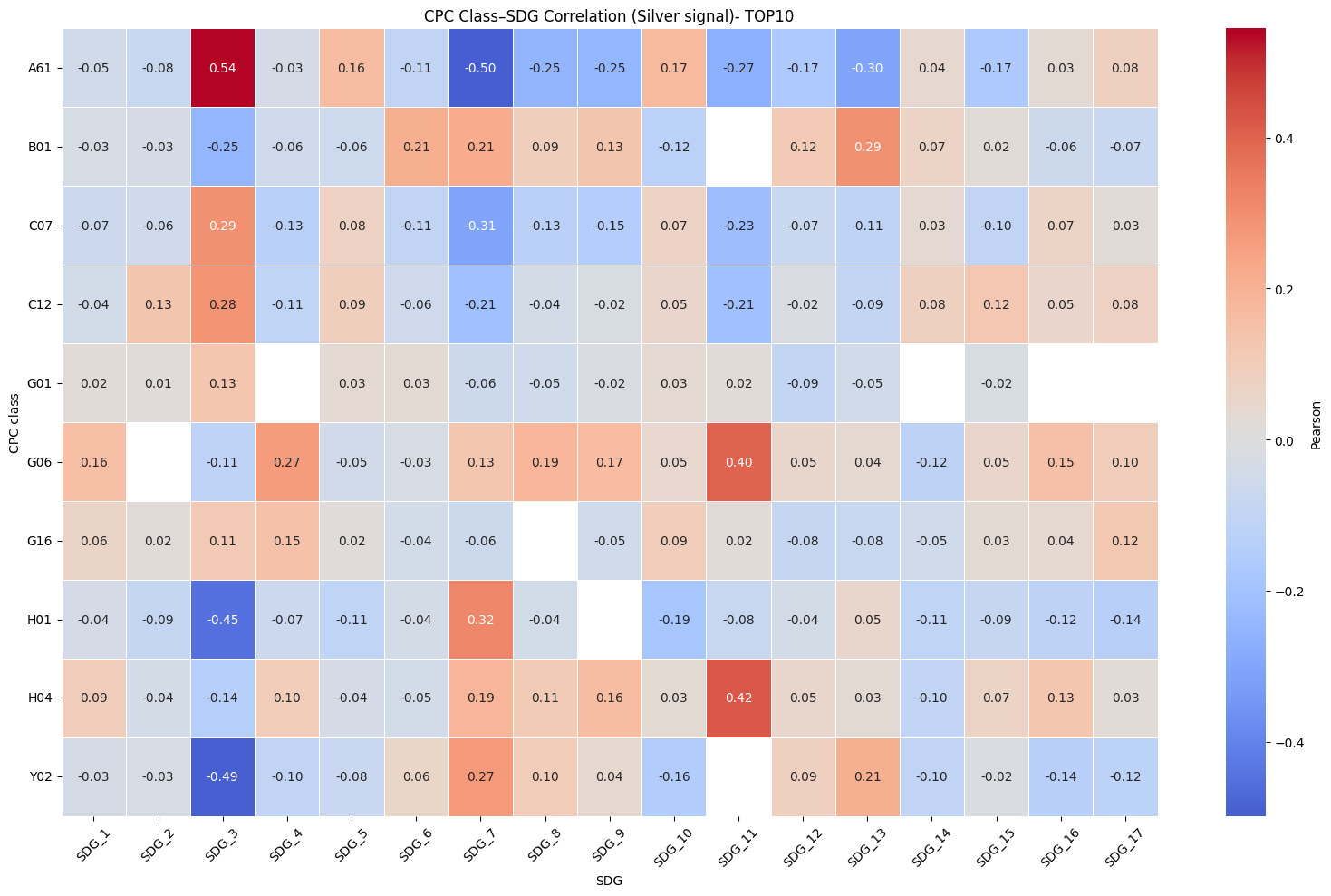}
        \caption{Pearson correlation between CPC classes and SDGs (Silver signal)}
        \label{fig:cpc_sdg_corr_silver}
    \end{subfigure}
    \caption{Correlation between CPC3 classes and SDG targets using two SDG labeling signals.}
    \label{fig:cpc_sdg_correlation_comparison}
\end{figure}

Moreover, we compare the distribution of SDG labels with the distribution of CPC codes. CPC codes are widely used for patent analytics and innovation studies due to their technical granularity and international consistency. Each patent can be assigned multiple CPC codes, which capture different aspects of its technological content.

In our dataset, we observe that SDG-relevant patents are broadly distributed across the technological landscape. Specifically, 124 out of the 137 three-digit CPC (CPC3) classes are present in our silver standard. This suggests that SDG-linked innovation is technologically diverse, although certain subclasses, such as \texttt{A61} (medical or veterinary science), \texttt{Y02} (climate change mitigation), and \texttt{C07} (organic chemistry), dominate in terms of volume and coverage.

We further explore the relationship between CPC classes and SDGs using Pearson correlation coefficients between CPC3 and SDG labels (both based on original NPL and a  silver signal). As shown in Figure~\ref{fig:cpc_sdg_corr_npl} and Figure~\ref{fig:cpc_sdg_corr_silver}, we find a range of statistically significant associations. For example, class \texttt{A61} shows a strong positive correlation with SDG~3 (Good Health), while \texttt{Y02} is positively associated with SDG~7 (Clean Energy) and SDG~13 (Climate Action). However, most relationships are moderate, highlighting the many-to-many mapping between technologies and societal goals.

\subsection{Downstream classification results} \label{sec:down}

The final step of our methodological pipeline involves assessing whether the silver-standard labels generated through weak supervision encode meaningful, learnable structure that can be effectively modeled from raw patent text. As discussed in Section~\ref{sec:class}, we conceptualize the SDG mapping task as a multi-output regression problem, where each patent is associated with a 17-dimensional vector representing continuous, non-exclusive alignment with each of the SDGs.

This formulation departs from traditional binary or multi-label classification approaches commonly used in prior SDG-related patent classification efforts \citep{maehara2025multi, hajikhani2022mapping}, and instead embraces a more realistic representation of sustainability-oriented innovation. A given patent may exhibit varying degrees of conceptual or technological relevance to multiple SDGs simultaneously. Hence, rather than enforcing hard class assignments, we opt for a regression-based strategy that preserves these nuances, aligning closely with the soft annotations produced by our labeling function.

To evaluate the learnability of the silver signal, we compare multiple supervised learning models trained on the silver-labeled dataset. All models are trained to predict the soft SDG distribution vectors using only patent title and abstract as input features. Specifically, we consider the following architectures: \texttt{BERT-for-patents} \citep{srebrovic2020leveraging}, \texttt{PAECTER},  \texttt{PAT-SPECTER} \citep{ghosh2024paecter} and \texttt{SCIBERT} \citep{beltagy2019scibert}. \footnote{The models were trained for 5 epochs with a batch size of 128, learning rate of $2 \times 10^{-5}$, and weight decay of 0.01. Input sequences were tokenized to a maximum of 128 tokens. The models were optimized with a weighted MSE loss function incorporating inverse-frequency SDG class weights. Training and evaluation were conducted on a single Tesla V100 GPU using \texttt{PyTorch}. Further, to address class imbalance, since certain SDGs are more prevalent than others, we also consider a \emph{weighted MSE loss}:
\begin{equation}
\mathrm{Weighted\ MSE} = \frac{1}{N} \sum_{p=1}^{N}\sum_{j=1}^{17} w_j \left(y_{pj} - \hat{y}_{pj}\right)^2
\end{equation}
where \(w_j\) is the weight for SDG \(j\), typically set inversely proportional to its frequency in the training data. This ensures that errors on underrepresented SDGs are penalized more heavily, improving learning across all classes.
}

\begin{table}[htbp]
\centering
\caption{Overall regression performance of different models on SDG prediction}
\label{tab:overall_mse_mae}
\begin{tabular}{lcc}
\toprule
\textbf{Model} & \textbf{MSE} & \textbf{MAE} \\
\midrule
BERT-for-patents         & 0.0196 & 0.0649 \\
PAECTER                  & 0.0188 & 0.0638 \\
\rowcolor{gray!15}
PAT-SPECTER              & 0.0187 & 0.0611 \\
SCIBERT                  & 0.0205 & 0.0634 \\
\bottomrule
\end{tabular}

\scriptsize
\emph{Notes:} Mean squared error (MSE) and mean absolute error (MAE) are computed over the 17-dimensional SDG regression targets for each patent. Lower values indicate better alignment between predicted and silver-standard SDG relevance scores.
\end{table}

Table~\ref{tab:overall_mse_mae} reports the regression performance of four language models trained to predict soft SDG relevance scores. Results are consistent across models, with mean squared errors ranging from 0.0187 to 0.0205 and mean absolute errors between 0.0611 and 0.0649. Among them, \texttt{PAT-SPECTER} achieves the best overall performance, obtaining both the lowest MSE and MAE. The convergence of results across architectures nonetheless indicates that the silver-standard SDG relevance scores generated via weak supervision encode robust, learnable patterns. While these aggregate metrics provide a useful overview, a more fine-grained analysis across the 17 SDGs reveals substantial heterogeneity in predictive difficulty, which we examine next.

\begin{table}[htbp]
\centering
\caption{Per-SDG regression performance (Mean Squared Error, MSE). Columns correspond to SDGs; rows are metrics.}
\label{tab:sdg_mse}

\begin{minipage}{\textwidth}
\centering
\subcaption*{SDG 1–9}
\begin{tabular}{lccccccccc}
\toprule
\textbf{Metric} & \textbf{SDG1} & \textbf{SDG2} & \textbf{SDG3} & \textbf{SDG4} & \textbf{SDG5} & \textbf{SDG6} & \textbf{SDG7} & \textbf{SDG8} & \textbf{SDG9} \\
\midrule
MSE & 3.35e-05 & 0.00651 & 0.15402 & 0.00121 & 0.00271 & 0.00614 & 0.06715 & 0.00097 & 0.05212 \\
\bottomrule
\end{tabular}
\end{minipage}

\vspace{0.8em}

\begin{minipage}{\textwidth}
\centering
\subcaption*{SDG 10–17}
\begin{tabular}{lcccccccc}
\toprule
\textbf{Metric} & \textbf{SDG10} & \textbf{SDG11} & \textbf{SDG12} & \textbf{SDG13} & \textbf{SDG14} & \textbf{SDG15} & \textbf{SDG16} & \textbf{SDG17} \\
\midrule
MSE & 0.00100 & 0.00678 & 0.00338 & 0.00981 & 0.00486 & 0.00052 & 0.00074 & 0.00072 \\
\bottomrule
\end{tabular}
\end{minipage}

\scriptsize
\justifying\emph{Notes:} Higher MSEs (e.g., SDG~3, SDG~7, SDG~9) reflect broader intra-class semantic variability; very low MSEs often correspond to rarer or more homogeneous goal profiles.
\end{table}

Table~\ref{tab:sdg_mse} reports per-SDG mean squared error. The per-SDG breakdown highlights a clear heterogeneity in predictive difficulty. SDG~3 (Good Health and Well-being), SDG~7 (Affordable and Clean Energy), and SDG~9 (Industry, Innovation and Infrastructure) show markedly higher error rates than the others. This is consistent with two well-known phenomena in machine learning: (\emph{i}) frequent labels with broad semantic coverage are harder to model due to their internal diversity (e.g., health-related patents span pharmaceuticals, devices, diagnostics, and healthcare systems, which increases noise in the regression target); and (\emph{ii}) imbalanced label distributions amplify error in the dominant classes while making rare classes appear artificially “easier” to predict. For example, the very low error observed for SDG~15 (Life on Land) or SDG~16 (Peace, Justice and Institutions) should not be interpreted as superior learnability, but rather as reflecting narrow and homogeneous subsets in the silver dataset.
\clearpage

\section{Limitations}
\label{sec:limitations}

This study proposes a weak supervision framework for mapping patents to SDGs in the absence of gold standard. While the approach demonstrates empirical validity and outperforms existing methods on several evaluation metrics, it is important to acknowledge several limitations that may affect its generalizability, reliability, and scope.

First, the construction of silver-standard labels relies exclusively on patents that cite SDG-tagged scientific publications, which introduces a form of selection bias in the training data. While the resulting classifier can, in principle, be applied to any patent, the labels used for training and validation are drawn only from a subset of patents with explicit NPL citations. This subset is not representative of the entire patent landscape: certain technological domains tend to cite scientific literature much more than others, such as life sciences \citep{jefferson2018mapping}. As a result, the model may be implicitly biased toward sectors with strong science–technology linkages, potentially underrepresenting SDG-relevant innovation in fields where NPL citations are rare or absent, therefore making certain societal challenges appear less relevant in technological landscapes than they truly are.This may affect both the balance of SDG coverage and the classifier’s ability to generalize across underrepresented domains.

Second, the evaluation of SDG label quality through structural metrics such as citation, inventor, and applicant homophily, though grounded in prior literature \citep[e.g.,][]{bergeaud2017classifying}, rests on heuristic proxies for semantic similarity. Shared inventors or applicants may indicate organizational or cognitive proximity but do not guarantee thematic alignment. Similarly, patent citations can reflect examiner practices or legal strategy rather than substantive knowledge flows \citep{alstott2017mapping}. As a result, high modularity scores provide evidence of internal coherence, but not definitive validation of label correctness.

Third, while the study adopts a multi-output regression approach to accommodate continuous, non-exclusive SDG relevance scores, it does not explicitly address the issue of class imbalance. Some SDGs are significantly more prevalent than others in the silver-standard dataset, which may bias model training and reduce sensitivity to underrepresented goals. The impact of this imbalance on prediction reliability, particularly for rare SDG classes, is not systematically investigated. Further work could explore advanced imbalance mitigation strategies and model calibration techniques to improve robustness across the full SDG spectrum.

Lastly, while the WS pipeline LLMs to abstract and align semantic content between patents and scientific literature, it remains sensitive to well-documented limitations of these models such as prompt variability, structural inconsistency, and domain-specific errors \citep{sclar2023quantifying, errica2024did}. These factors can influence the quality and stability of the extracted concepts that underpin the labeling function. However, the use of aggregation and ranking strategies partially mitigates these risks by reducing the impact of outlier outputs and promoting consensus-driven alignment \citep{dey2025uncertainty}. Further, the semantic extraction process relies on a fixed ontology of \textit{functions}, \textit{solutions} \citep{zhai2022patent}, and \textit{applications}, which may omit context-specific or cross-cutting dimensions present in patent texts. This constraint can limit the flexibility of the labeling function and potentially overlook innovations that do not neatly conform to these predefined categories.

\clearpage

\section{Conclusion}\label{sec:conclusion}

This study addresses a central challenge in mapping patents to the SDGs: the absence of a large-scale, publicly available gold standard for supervised learning. In the absence of authoritative labels, we treat NPL citations to SDG-tagged scientific publications as informative but incomplete supervision anchors. Although NPL citations are known to reflect conceptual proximity between science and technology, they are systematically noisy and partial \citep{narin1997increasing, he2007evidence, nagaoka2015use, callaert2006traces}. To extract broader and more meaningful SDG signals from patent text, we apply weak supervision, a family of methods that enables programmatic label generation through the design of heuristic labeling functions, rather than manual annotation \citep{Ratner_2017, cohen2019interactive}.

In particular, we introduce a labeling function that combines LLM-guided semantic concept extraction with embedding-based document alignment. Specifically, we use prompting strategies to extract structured semantic elements, \textit{functions}, \textit{solutions}, and \textit{applications}, from both patents and scientific literature. These structured elements are compared across domains using vector similarity and ranking methods to estimate SDG relevance for each patent. The resulting dataset assigns soft, multi-label SDG scores to more than 30,000 patents, with no need for hand labeling or domain-specific fine-tuning.

Empirical evaluation proceeds in two stages. First, we show that the proposed LF outperforms a wide range of encoding systems in reproducing SDG labels inferred from NPL citations. The method demonstrates higher recall and better coverage across both frequent and rare SDG classes. Second, we validate the plausibility of the extra NPL generated labels through network theory. Using overlapping modularity \citep{nicosia2009extending}, we examine whether patents with similar SDG profiles cluster more tightly in citation networks (thematic similarity), inventor networks (cognitive similarity), and applicant networks (organizational similarity). In all three cases, our method yields stronger modularity scores than CPC-derived labels or NPL-only proxies, suggesting that the generated labels are more consistent with real-world knowledge structures \citep{bergeaud2017classifying, arts2018text, yoo2023novel}.

This study contributes to the literature in several ways. First, it shows that semantic methods only partially replicate the SDG signal derived from NPL, but can meaningfully enrich it by recovering additional sound patent–SDG associations. Second, it demonstrates that text-based classification methods can produce more relevant groupings than traditional technological taxonomies such as the CPC system, particularly in policy-relevant or interdisciplinary domains such as the SDGs \citep{bergeaud2017classifying}. Third, it highlights that LLMs may underperform in zero-shot classification of patents to SDGs, but excel in extracting structured semantic elements in few-shot or prompt-guided settings. Rather than using LLMs for direct labeling, our approach leverages them as high-recall tools for semantic abstraction, providing domain-relevant features that improve downstream learning. This aligns with recent work showing the advantages of prompt engineering and structured output extraction over raw zero-shot prediction in domain-specific applications \citep{agrawal2022large, dagdelen2024structured, yoshikawa2025large}.

Future research should aim to diversify sources of weak supervision by integrating expert-labeled data, policy documents, or crowd-sourced validation. Introducing human-in-the-loop feedback would also help improve label interpretability and trust. Furthermore, addressing class imbalance through data augmentation, label smoothing, or reweighting strategies could enhance model robustness across the full SDG space. Finally, extending the method to multilingual patent corpora and related classification schemes, such as environmental, social, and governance (ESG) frameworks or climate technology taxonomies, would broaden its relevance for global innovation monitoring and science-policy evaluation.

In sum, this work demonstrates that semantically grounded, weakly supervised methods offer a scalable, interpretable, and empirically validated pathway to link patents to the SDGs, bridging the gap between innovation metrics and societal impact.

\clearpage

\appendix

\newgeometry{top=1.5cm, bottom=1cm, left=1cm, right=1cm}

\section{Appendix} \label{app:1}

\subsection{SDG queries splitting methodology} \label{a1}

This Section provides detail the method used to split the long and complex Boolean SDG queries of the Elsevier SDG Research Mapping Initiative that allows users to identify scientific publications that are relevant to the 17 SDGs. The process ensures that logically equivalent subqueries are generated within system-imposed character limits (e.g., 2000 characters), while preserving the logical structure of the original query.
 
 To this end, we propose a Boolean query splitting and optimization framework that leverages the formal properties of Boolean algebra to ensure both computational scalability and logical fidelity.

The process begins with a rigorous sanitization and tokenization step, where raw query strings are normalized to ensure consistent syntax. This includes enforcing quotation of search field elements (e.g., transforming \texttt{AUTHKEY(rural)} to \texttt{AUTHKEY("rural")}) and preparing the query for parsing. The tokenized string is then parsed into a hierarchical tree representation, where each node explicitly captures the logical relationships (AND, OR, NOT, AND NOT) and parenthetical nesting defined in the original query.

Central to our approach is the decision to use the \texttt{OR} operator as the primary axis for splitting. This choice is justified both theoretically and practically: the commutative and associative properties of disjunction allow any disjunctive block (e.g., $A \lor B \lor C$) to be partitioned into smaller, independently executable subqueries whose result sets can be safely recombined via set union, without altering the final logical outcome. This is not generally possible with \texttt{AND}-based conjunctions or \texttt{AND NOT} exclusions, which impose stronger dependencies between terms.

For cases involving exclusion clauses — queries of the form $(A \lor B) \land \lnot (C \lor D)$ — we apply distributive expansion, generating all permutations such as $(A \land \lnot C)$, $(A \land \lnot D)$, $(B \land \lnot C)$, $(B \land \lnot D)$. This ensures that the effect of the exclusion is correctly propagated across all positive blocks, preserving the semantics of the original query even under decomposition.

The system is further designed to handle long conjunctions (i.e., \texttt{AND}-separated positive chains) by batching or further splitting them when they exceed configured size thresholds, ensuring that no subquery violates the API constraints. Importantly, the framework includes recursive mechanisms: at each node of the Boolean tree, it evaluates whether local splitting, batching, or deeper recursion offers the most efficient and logically sound partitioning, balancing computational cost against semantic accuracy.

Overall, our method provides a robust, general-purpose solution for executing large and complex Boolean queries, with particular relevance for SDG-related bibliometric tasks, but equally applicable to any domain where search queries approach the structural or operational limits of available retrieval systems.

\clearpage

\subsection{Distribution of SDG papers over time} \label{a1b}

\begin{figure}[h]
\centering
\includegraphics[scale=0.45]{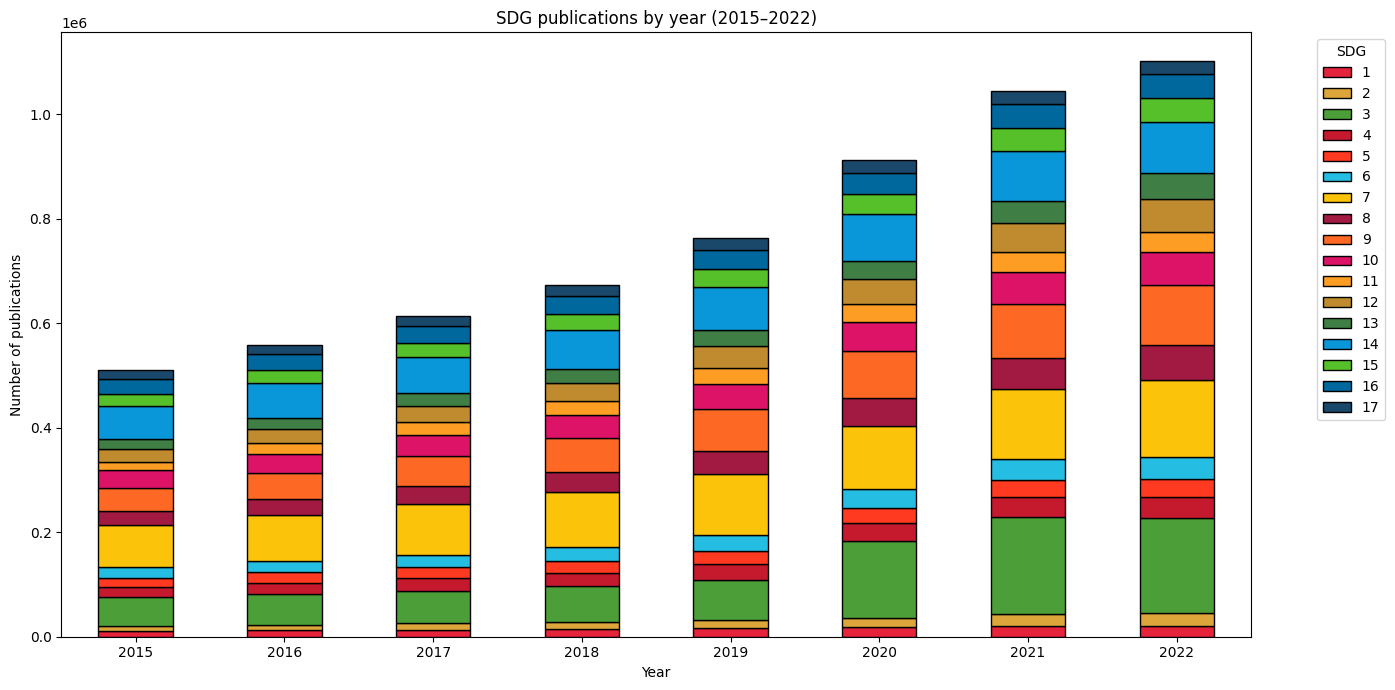}
\caption{SDG publications trend (2015--2022)}
\label{sdgpub}
\scriptsize
\justifying\emph{Notes:} The Figure displays the number of SDG-tagged papers published annually between 2005 and 2022 by the Elsevier project. If a paper is classified under multiple SDGs, it is counted once per SDG. Only English-language journal articles are included.
\end{figure}

\clearpage
\newpage

\newpage
\newgeometry{left=2cm, right=2cm, top=3cm, bottom=3cm, footskip=0.5cm}
\bibliographystyle{elsarticle-num-names}

\clearpage

\bibliography{biblio}

\end{document}